\documentclass[runningheads]{llncs}

\usepackage{graphicx}
\usepackage{comment}

\usepackage{amsmath,amssymb}
\usepackage{color}
\usepackage{url}
\usepackage{hyperref}
\usepackage{amsthm}
\usepackage{booktabs}
\usepackage{float} 
\usepackage{subcaption}
\usepackage{listings}
\usepackage{xcolor}
\usepackage{wrapfig}
\usepackage{xr}

\newif\ifreview
\reviewtrue
\reviewfalse

\ifreview
	\usepackage{lineno}

	\linenumbers
\fi

\newcommand{\domain}{\ensuremath{\textbf{D}}}
\newcommand{\domaind}{\ensuremath{p_\mathbf{D}}}
\newcommand{\domaindi}{\ensuremath{p_\mathbf{D_i}}}
\newcommand{\sdomain}{\ensuremath{\hat{\textbf{D}}}}
\newcommand{\mean}[2]{\mathbb{E}_{#1}\left[ #2 \right]}
\newcommand{\variance}[2]{\text{Var}_{#1}\left[ #2 \right]}

\newcommand{\sotheta}{\ensuremath{\hat{\boldsymbol{\theta}}}}
\newcommand{\otheta}{\ensuremath{\boldsymbol{\theta}}}

\newcommand{\stream}{\ensuremath{\textbf{S}}}
\newcommand{\streams}{\ensuremath{\hat{\textbf{S}}}}
\newcommand{\mypara}[1]{\paragraph{#1}\mbox{}}

\lstdefinestyle{mypython}{
  language=Python,
  basicstyle=\ttfamily\footnotesize, 
  keywordstyle=\color{blue},
  commentstyle=\color{gray},
  stringstyle=\color{red},
  showstringspaces=false,
  breaklines=true,
  frame=single,
  captionpos=b
}

\begin{document}


\def\SubNumber{99}

\def\GCPRTrack{Main Track}

\title{On the Dangers of Bootstrapping Generation for Continual Learning and Beyond}

\ifreview
\else

	\author{Daniil Zverev\inst{1} \and
	A. Sophia Koepke\inst{1,2}\and
	Joao F. Henriques\inst{3}}
	
	\authorrunning{D. Zverev et al.}
	
	\institute{Technical University of Munich, MCML \and University of Tübingen, Tübingen AI Center \and  University of Oxford}
\fi

\maketitle              

\begin{abstract}
The use of synthetically generated data for training models is becoming a common practice. 
While generated data can augment the training data, repeated training on synthetic data raises concerns about distribution drift and degradation of performance due to contamination of the dataset. 
We investigate the consequences of this bootstrapping process through the lens of continual learning, drawing a connection to Generative Experience Replay (GER) methods. We present a statistical analysis showing that synthetic data introduces significant bias and variance into training objectives, weakening the reliability of maximum likelihood estimation. We provide empirical evidence showing that popular generative models collapse under repeated training with synthetic data. We quantify this degradation and show that state-of-the-art GER methods fail to maintain alignment in the latent space. Our findings raise critical concerns about the use of synthetic data in continual learning.

\end{abstract}

\section{Introduction}
\label{sec:intro}

Generative models have become essential for modern machine learning, being used in various tasks ranging from text generation to image synthesis. These models, such as GPT-based large language models~\cite{openai2023gpt4} and diffusion-based image generators like Midjourney, are now key components in consumer and industrial applications. A natural consequence of their proliferation is the growing presence of synthetic data in the publicly available data corpus~\cite{cui2023said}. As this trend continues, future models are likely to be trained on data that was itself generated by other models.
This growing reliance on synthetic data raises questions about the consequences of repeatedly training models on data generated by earlier models. While synthetic data can temporarily enrich datasets, incorporating generated samples into future training regimes risks long-term degradation of model performance due to distributional drift and statistical contamination.

This phenomenon is closely related to continual learning (CL), specifically in the form of Generative Experience Replay~\cite{shin2017continual,gao2023ddgr}. In GER, a model is exposed to a stream of non-i.i.d.\ tasks and maintains performance across them by using a generative model to replay synthetic samples from past tasks. This setup reflects broader trends in machine learning, where synthetic data is reused across training cycles.

In this paper, we study the statistical and empirical consequences of this synthetic bootstrapping loop. We begin by formalising the continual learning setup and examining the statistical errors introduced by synthetic data. In particular, we consider bias and variance in maximum likelihood estimators when real data is replaced by generated samples. We then analyse GER continual learning algorithms, identifying how these statistical errors manifest in state-of-the-art methods. Our experiments highlight that generative models exhibit instability when repeatedly trained on synthetic samples. We provide empirical evidence that, over time, synthetic datasets diverge from their original distributions, leading to a degradation in downstream performance and increased divergence in latent space representations.

To summarise, our contributions are as follows:
\begin{enumerate}
\item We provide a theoretical analysis demonstrating how repeated training on synthetic data introduces bias and variance into standard training objectives, weakening the statistical guarantees of generative model learning.
\item We perform controlled experiments on GANs and diffusion models, empirically showing that repeatedly training on generated data leads to distributional drift and downstream performance degradation, even under ideal conditions.
\item We quantify the divergence of synthetic and real data and show that state-of-the-art GER methods fail to prevent latent space separation between the two domains.
\end{enumerate}

Our findings provide a cautionary perspective on synthetic data usage, along with theoretical grounding for understanding the limitations of GER in continual learning.

\section{Related work}
\paragraph{Continual learning and generative replay.}
CL addresses the challenge of training models on non-stationary data distributions without the important problem of catastrophic forgetting~\cite{grossberg1982does,prabhuGDumbSimpleApproach2020,rebuffiICaRLIncrementalClassifier2017,shin2017continual,lesortContinualLearningRobotics2019}. This is commonly addressed by revisiting old data through experience replay methods~\cite{rolnick2019experience,lopez2017gradient,rebuffiICaRLIncrementalClassifier2017,chaudhry2019tiny}. A prominent subfamily of methods, Generative Experience Replay (GER) \cite{shin2017continual,DBLP:journals/corr/abs-1711-10563,DBLP:journals/corr/abs-1809-10635,robins1995catastrophic}, mitigates forgetting by using generative models to recreate past task data~\cite{shin2017continual}. GAN Memory~\cite{cong2020gan} and DDGR~\cite{gao2023ddgr} further scale GER to more complex datasets using GANs and diffusion models respectively. Our work complements these efforts by showing that even state-of-the-art GER systems suffer from systematic degradation when synthetic data is bootstrapped over time. We quantify this with both statistical analysis and empirical metrics, revealing that synthetic data contamination in GER setups leads to increasing domain shift, classifier collapse, and misalignment in latent space.

\mypara{Synthetic data for training.}
The rise of generative models has led to an increasing amount of synthetic content in public datasets. For instance, \cite{matatov2024examining} estimated that 0.2\% of Reddit posts are AI-generated, while \cite{cui2023said} found that annotators can distinguish synthetic from real online data with 96.5\% accuracy, revealing a measurable domain shift. According to \cite{villalobos2022will}, we may soon exhaust high-quality human-generated data, necessitating reliance on synthetic data to train future models. 
Several recent works have observed performance degradation when synthetic data is introduced into training~\cite{alemohammad2024selfconsuming,hataya2023will,shumailov2023curse,shumailov2024collapse}. Specifically, \cite{hataya2023will} investigated the use of synthetic samples for ImageNet~\cite{deng2009imagenet} and COCO~\cite{lin2014microsoft} categories, showing notable drops in classification performance. However, these works make strong theoretical assumption about generative models, modeling them as a Gaussian distributions, our work, in contrast, analyses how replacing real data with synthetic samples impacts the statistical assumptions underlying maximum likelihood training, offering a principled explanation for the observed empirical failures, applicable to any generative model.

\mypara{Measuring domain shift.} The broader problem of domain shift~\cite{kouw2018introduction} and out-of-distribution generalisation~\cite{yang2024generalized} is well-studied, particularly in transfer learning and robustness contexts. Metrics such as FID~\cite{heusel2017gans}, OTDD~\cite{alvarez2020geometric}, have been proposed to measure divergence between datasets. We adopt and extend these tools to measure how synthetic data diverges from real distributions over time in a continual training loop. We evaluate synthetic collapse not just as a static phenomenon but as a gradual process, showing how recursive training causes statistical instability and undermines the training objectives.

\section{Continual Learning}
\label{sec:continual_learning}


In offline continual learning, a model is trained sequentially on a stream of domains $\stream  = \{\domain_i\}_{i=1}^T$. The entire training process is composed of multiple training and evaluation phases which are commonly referred to as domains in the Domain Adaptation literature \cite{wang2018deep} and as tasks in Continual Learning. In this work, each task contains one domain. Thus, the terms become interchangeable. During each task, we only have access to one domain $\domain_t$ for training, and we evaluate on the union of all previous domains $\domain_{\textbf{val}} = \bigcup_{i=1}^{t-1} \domain_i$.
In the following, we define key terminology for the continual learning process. In particular, we define domain and memory, used for retaining information across tasks.

\begin{definition}[Domain] A domain $\domain_i$ consists of input-label pairs $\{x_j, y_j\}_{j=1}^{N_i}$. 
If the training objective uses unlabelled data, $y_j=\emptyset$. Each sample within a domain is distributed according to the domain distribution $\domaindi$.
\end{definition}


\begin{definition}[Memory] The memory contains information that is preserved between tasks. We denote it as $\mathcal{M}_{t-1}$. It can contain any parameters $\theta$ necessary for training and inference, such as model weights, regularisation parameters, or samples from the previous domain.
\end{definition}


\begin{definition}[Continual learning process]
  The continual learning process consists of an operator $P$ sequentially applied $T$ times on the initial memory $M_0$,
  as $\mathcal{M}_t = P(\domain_t, \mathcal{M}_{t-1})$ for $t=1,\ldots,T$.
\end{definition}
A connection between continual learning and the standard training of large models (e.g.\ those in the GPT family \cite{radford2018improving}) can be established by considering assumptions about data 
distribution shifts across domains. 
\begin{wraptable}{r}{0.45\textwidth}
  \centering
  \resizebox{0.44\textwidth}{!}{
  \begin{tabular}{ccl}
    \toprule
     \textbf{Release date} & \textbf{Task} & \textbf{Domain} \\
    \midrule
     2024-05-13 & 1 & $\domain_1 = \text{Initial domain}$ \\
     2024-08-06 & 2 & $\domain_2 = \text{New data} + \textit{S}(\domain_1)$ \\
     2024-09-03 & 3 & $\domain_3 = \text{New data} + \textit{S}(\domain_2)$ \\
    \bottomrule
  \end{tabular}
  }
  \caption{GPT-4o release cycle. $\textit{S}$ denotes the subset of a domain used to train.}
  \label{tab:lgm_release_cycle}
\end{wraptable}
When training large models, domains are non-disjoint, i.e.\ the same data instance $x_k$ can appear in multiple tasks and their related domains. 
As shown in Tab.~\ref{tab:lgm_release_cycle}, a task corresponds to the time between two public checkpoint releases for GPT-4o. The task domain is the data used for training the corresponding model weights checkpoint.
Furthermore, the memory $\mathcal{M}_{t}$ keeps only parameters necessary for training/inference and the process $P$ at each task consists of SGD training over the corresponding domain $\domain_t$. 


\subsection{Synthetic data error}
\label{sec:stat_errors}

When training neural networks, unavoidable errors occur. The most fundamental is the model family limitation $\textbf{ERROR}_{\text{model}}$, in statistical learning theory \cite{bousquet2003introduction} commonly referred to as ``approximation error''. Every architecture is parametrised by a set of parameters $\theta$ and has inherent expressiveness limitations. Over time, the performance of models increases \cite{kaplan2020scaling}, with new architectures like Transformers \cite{vaswani2017attention} overthrowing LSTMs \cite{hochreiter1997long}, and Diffusion models \cite{rombach2021highresolution} surpassing GANs \cite{goodfellow2014generative} on benchmarks, thereby reducing $\textbf{ERROR}_{\text{model}}$. 

In addition to model limitations, another statistical error, $\textbf{ERROR}_{\text{synth}}$, arises when parts of the data stream $\stream$ are replaced with synthetic, artificially generated domains $\sdomain$. The distributional differences between $\sdomain$ and the original domain $\domain$ can adversely affect the validation loss when training on one domain and validating on another. This issue is particularly significant when there is no way of knowing whether a dataset was contaminated. For instance, publicly available internet data used to train models like GPT-4 \cite{openai2023gpt4} now contains substantial generated content, a trend expected to continue \cite{villalobos2022will}.

Synthetic domains may differ from real domains for several reasons. We argue that the primary cause is the inevitable $\textbf{ERROR}_{\text{model}}$ of the generative model used to produce the synthetic domain. 
Another reason could be undersampling, where $|\sdomain| \ll |\domain|$, leading to a loss of information.
Both are of great importance. This work provides theoretical proofs and practical examples demonstrating why $\textbf{ERROR}_{\text{synth}}$ must be addressed.  

\subsection{Training objective}\label{sec:train_objective}
For large generative models such as LLMS \cite{naveed2023comprehensive}, GANS, and Diffusion models, the training objective is commonly based on the Maximum Likelihood Estimator (MLE):
%
\begin{equation}
\label{eq:mle}
    \max_\theta L(\stream, \theta) = \ln p_\theta(\domain_1,..., \domain_t) =  \sum_{i=1}^{t}\sum_{j=1}^{N_i} \ln p_\theta(x_{ij}),
\end{equation}
with the training data stream up to phase $t$. The term $p_\theta(\domain_1, \ldots, \domain_t)$ denotes the likelihood of the data given the model parameters $\theta$.

To understand the impact of synthetic data on the estimation process and model performance, we analyse the variance and bias of the MLE under a constantly changing distribution $\domain_t$. Each change corresponds to the introduction of synthetic data.
\begin{figure*}[t]
  \centering
  \begin{subfigure}[t]{0.28\textwidth}
    \centering
    \includegraphics[width=\linewidth]{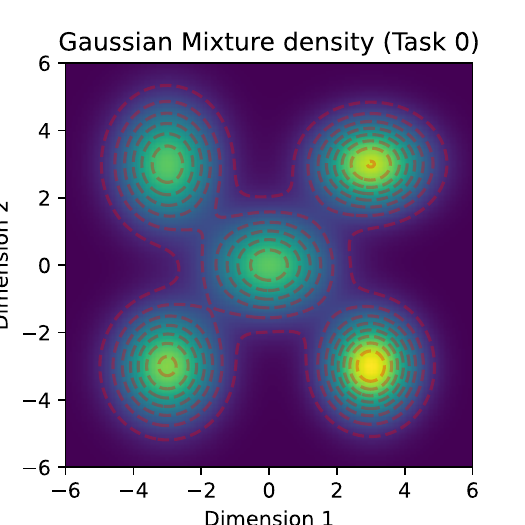}
    \caption{Density function of a Mixture of 5 Gaussians (GMM), representing $p_{\otheta}$.}
    \label{fig:density_phase_0}
  \end{subfigure}
  \hspace{0.01\textwidth}
  \begin{subfigure}[t]{0.28\textwidth}
    \centering
    \includegraphics[width=\linewidth]{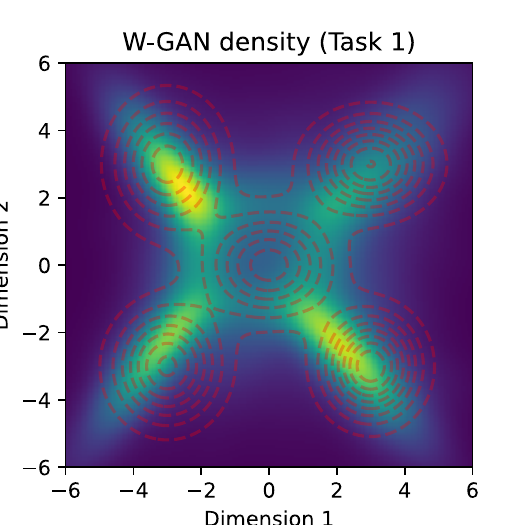}
    \caption{WGAN trained on GMM samples, representing first task distribution $p_{\sotheta_1}$.}
    \label{fig:density_phase_1}
  \end{subfigure}
  \hspace{0.01\textwidth}
  \begin{subfigure}[t]{0.33\textwidth}
    \centering
    \includegraphics[width=\linewidth]{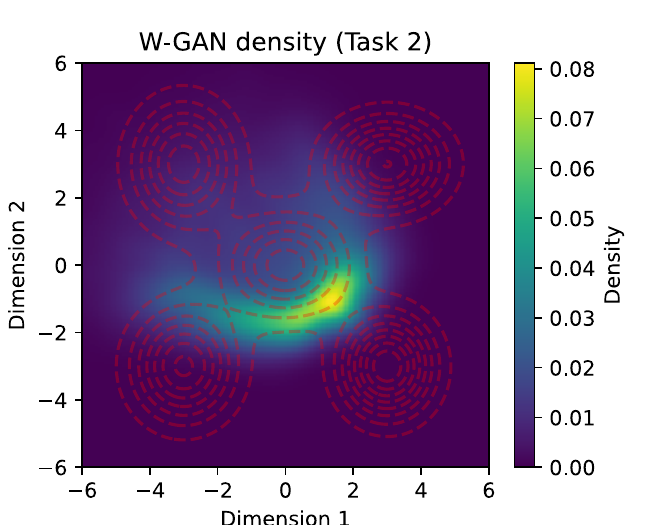}
    \caption{WGAN trained on $p_{\sotheta_1}$ samples, representing second task distribution $p_{\sotheta_2}$.}
    \label{fig:density_phase_2}
  \end{subfigure}

  \caption{Illustration of three training phases of a WGAN \cite{arjovsky2017wasserstein}. Datasets are undersampled as described in Sec.~\ref{sec:stat_errors}, resulting in a domain shift between synthetic and real data. Densities are estimated using kernel density estimation (KDE) \cite{scott2015multivariate}.}
  \label{fig:gan_density_phases}
\end{figure*}
Without loss of generality and for notational simplicity, we consider in the following the case where the entire domain $\domain_1$ is replaced by a synthetic domain $\sdomain_1$.

\mypara{Bias.} 
The generative model is trained on a mix of synthetic and non-synthetic domains with the objective of maximising the MLE:
\begin{align}
\label{eq:mixed-mle}
    \max_{\hat{\theta}} L(\streams, \hat{\theta}) &= \ln p_{\hat{\theta}}(\sdomain_1,\domain_2..., \domain_t).
\end{align}
Here, $\sotheta$ corresponds to the set of weights obtained after optimising Eq.~\ref{eq:mixed-mle}, and $\otheta$ are the weights obtained from Eq.~\ref{eq:mle}. In practice, the learned distribution $p_{\sotheta}$ assigns non-zero probabilities to samples from $\streams$ and near-zero probabilities to samples not covered by $\streams$.

For synthetic data, if $\domain_1$ and $\sdomain_1$ differ, $p_{\sotheta}$ will assign low probabilities to samples from $\domain_1$ and high probabilities to samples from $\sdomain_1$. This discrepancy creates a significant statistical problem, making the MLE estimator biased with respect to the original optimisation maximum,
since the validation loss is evaluated as an expectation over the real distribution $\domaind$.
Using the properties of $\ln$, we can write the bias between the estimators as:
\begin{equation}
\label{eq:mle-bias}
    \textbf{bias}(L(\stream, \sotheta), L(\stream, \otheta)) 
    = \mathbb{E}_{\domaind}\left[ L(\stream, \sotheta) - L(\stream, \otheta) \right] \nonumber
    = \mathbb{E}_{\domaind}\left[ \ln \frac{p_{\sotheta}(\domain_1)}{p_{\otheta}(\domain_1)} \right] .
\end{equation}
%
%
This shows that as the synthetic domain $\sdomain_1$ shifts further from the original domain $\domain_1$, the value of $p_{\sotheta}$ on $\domain_1$ decreases relative to the fixed $p_{\otheta}$, and the bias increases.

We demonstrate this empirically in Fig.~\ref{fig:gan_density_phases}, showing the difference between density values in the case of 5 Gaussians distributed over a 2-dimensional space (see Supplementary for more details). The distribution learned by a Wasserstein GAN (WGAN) \cite{arjovsky2017wasserstein} deviates from the target distribution $p_{\otheta}$, which increases the bias as described in Eq.~\ref{eq:mle-bias}.

The bias between the two MLEs reduces the statistical reliability of the obtained model and undermines the alignment between the training and validation objectives. 
By inserting synthetic data, we weaken the statistical guarantees for the validation scores on real domains.

\mypara{Variance.} 
We compute the variance over the real domain distribution $\domaind$ as
\begin{equation}
    \variance{\domaind}{L(\stream, \sotheta)} =  \variance{\domaind}{\sum_{i=1}^t \ln p_{\sotheta}(\domain_i)}.
\end{equation}
Similar to the bias, the problem with the variance arises on the replaced domain $\domain_1$ where $p_{\sotheta}$ returns low values. 
The variance can be expanded as:
\begin{equation}
\label{eq:mle-variance}
    \variance{\ensuremath{p_{\mathbf{D}}}}{L(\stream, \sotheta)} =  \mean{\domaind}{\ln p_{\sotheta}(\domain_1)^2} - \mean{\domaind}{\ln p_{\sotheta}(\domain_1)}^2 + \nu,
\end{equation}
where $\nu$ is the variance of the MLE on non-replaced domains.
\begin{wrapfigure}{r}{0.5\textwidth}
  \centering
  \begin{minipage}[t]{0.48\linewidth}
    \centering
    \includegraphics[height=3cm]{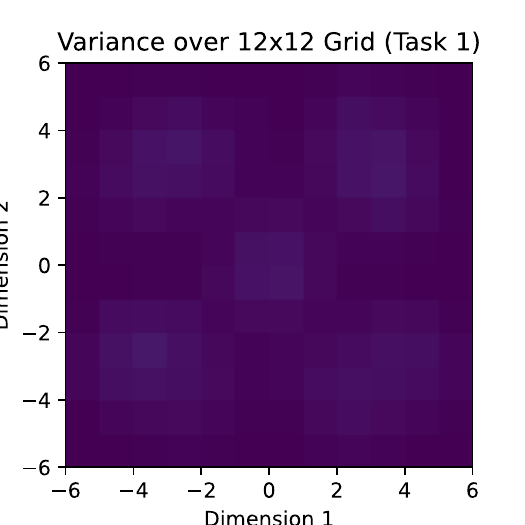}
    
    {\footnotesize (a) Variance after task $p_{\sotheta_1}$.}
  \end{minipage}%
  \hfill
  \begin{minipage}[t]{0.48\linewidth}
    \centering
    \includegraphics[height=3cm]{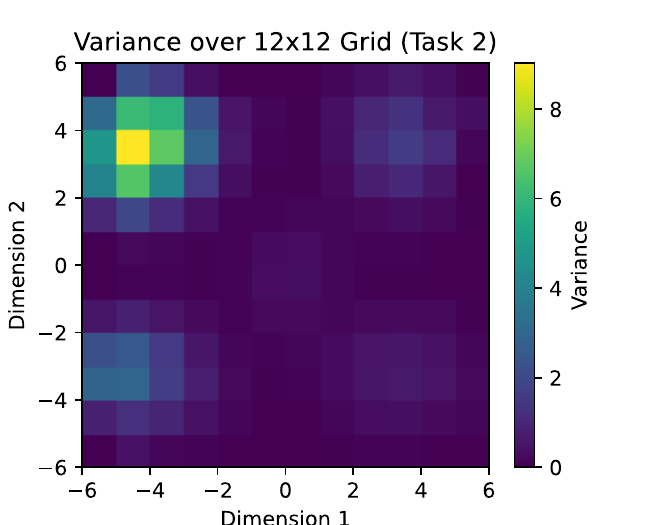}
    
    {\footnotesize (b) Variance after task $p_{\sotheta_2}$.}
  \end{minipage}
  \caption{Comparison of variances.}
  \label{fig:gmm_vs_gan_density}
  \vspace{-3em}
\end{wrapfigure}
Smaller $p_{\sotheta}$ on  $\domain_1$ leads to higher estimator variance as the first term in Eq.~\ref{eq:mle-variance} is larger than the second.

As illustrated in Fig.~\ref{fig:gmm_vs_gan_density}, the variance is high in subdomains where $p_{\sotheta}$ is low but $p_{\theta}$ is high (see Fig.~\ref{fig:gan_density_phases}), and lower where both distributions intersect. 
This demonstrates an additional statistical challenge: in regions where the domains $\domain_1$ and $\sdomain_1$ diverge, the MLE becomes not only biased but also exhibits high variance. This further undermines the reliability of the obtained weights $\sotheta$ when validated on real domains.
The presence of non-zero bias and high variance undermines the core optimisation assumptions of generative models.

\section{Continual Learning methods and synthetic data error}

In continual learning, we have a non-i.i.d. setup where we have access only to the current domain $\domain_t$. The previous domains $\{\domain_i\}_{i=1}^{t-1}$ are commonly replaced with a memory buffer~\cite{wang2023comprehensive}. However, we still validate our model on the entire data stream $\stream$. This leads to two objective functions: the training objective $L_t(\domain_t, \mathcal{M}_{t-1})$, and the validation objective $L_{\text{val}}(\domain_1, \ldots, \domain_{t})$.

The discrepancy between the memory $\mathcal{M}_{t-1}$ and the past domains $\{\domain_i\}_{i=1}^{t-1}$ increases the difference between the minima of the two objectives. Over the past years, the continual learning community has proposed numerous architectures to address this discrepancy \cite{wang2023comprehensive,rebuffiICaRLIncrementalClassifier2017,chaudhry2019tiny,lopez2017gradient}.
In the following sections, we will show that even state-of-the-art models in generative continual learning cannot mitigate $\textbf{ERROR}_{\text{synth}}$.

\subsection{Generative Experience Replay}
Following our notation from Sec.~\ref{sec:continual_learning}, Generative Experience Replay (GER) methods~\cite{shin2017continual} are trained with the MLE objective in the Continual Learning process as follows:
\begin{align}
\label{eq:ger_process}
    &\max_\theta L(\streams, \theta) = \ln p_\theta(\sdomain_1,\cdots, \sdomain_{t-1},\domain_t), \\ 
    &\mathcal{M}_1 = P(\domain_1), \, \cdots \, ,\mathcal{M}_T = P(\domain_T, \sdomain_{T-1},\cdots, \sdomain_1).
\end{align}
GER methods use a generative model to approximate samples from past domains. They offer an efficient trade-off between the number of generated samples and the memory required to store model parameters. However, they are harder to train as catastrophic forgetting can occur in both the downstream and generative models. GER methods can be divided into two categories: those that train the generative model simultaneously with the downstream model, and those that train them separately (visualised in the Supplementary).

\mypara{Joint training.}
For joint training the generative and downstream models are trained simultaneously and can be separate models \cite{kemker2017fearnet,gao2023ddgr} or one model \cite{van2018generative}. In this case, the losses are calculated on the same batch for both models, and weights are updated synchronously. 
At the end of task $t$, we utilise the generator to sample past domains $\{\sdomain_i\}_{i=1}^t$, following Eq.~\ref{eq:ger_process}.

\mypara{Separate training.}
For separate training \cite{shin2017continual,cong2020gan}, the downstream model is decoupled from the generative model. First, the generative model is trained on the real domain $\domain_t$. Then, a generated synthetic training dataset $\sdomain_t = {(x_i, y_i)}$ is used for the downstream models. In a subsequent task, the next generative model is trained using both the new task's $\domain_{t+1}$ training samples and the synthetically generated dataset $\sdomain_t$ from previous tasks. This leads to decorrelated gradients, as the loss is calculated on different batches.

\subsection{Quantitative analysis of the stability of generative models}\label{sec:quantitative_analysis_gen}
In both GER training setups, the model is exposed to both real and synthetic domains, with the synthetic data $\sdomain$ only containing information about past domains. 
Eqs.~\ref{eq:mle-bias} and \ref{eq:mle-variance} demonstrate the relevance of the statistical difference between the real and synthetic domains $\domain$ and $\sdomain$, respectively. We quantify this using the following metrics.

\mypara{Fréchet Inception Distance} (FID)~\cite{heusel2017gans} measures the Fréchet distance $d_{F}$ between two feature distributions extracted from images using the InceptionNet model \cite{DBLP:journals/corr/SzegedyVISW15}. We compute two scores. For class-unconditional generative models, we compute the standard FID score. For class-conditioned models, we introduce a Conditional FID which computes the distance for each class separately (see Supplementary):
\begin{equation}
    \mean{Y}{d_F(X_1|Y , X_2|Y)^2} = \frac{1}{k} \sum_{i=1}^k d_F(X_1 |\, Y = i, X_2 |\, Y = i)^2.
\end{equation}
%
Lower FID scores indicate that the generated images are closer to the real image distribution, making it a suitable metric for assessing divergence between $\domain_{\text{dataset}}$ and  $\sdomain_{\text{dataset}}$.

\mypara{Optimal Transport Dataset Distance} (OTDD)~\cite{alvarez2020geometric} is a model-agnostic metric introduced to measure the distance between datasets, in our case real and synthetic, in a way that does not depend on training a model or having matching label sets. OTDD computes distances between feature-label pairs across domains by combining geometric awareness with Wasserstein distances, which aligns datasets based on both features and labels. 

\mypara{Classification degradation}
measures the stability of the generative model by considering the classification accuracy. For every class-conditioned generative model and each task $t$, a ResNet18 classifier \cite{he2015deep} is trained on the synthetic dataset $\sdomain_t$ with labels $\hat{y}$. The classifier is then validated on the real dataset $\domain_{\text{dataset}}$, and the best validation score per task is reported.

While FID and OTDD are proxy metrics that measure the divergence of the synthetic data $\hat{x}$ from the real domain, the classification score provides a direct measure of how accurately the synthetic classes $\hat{y}$ represent the real distribution. A drop in classification performance suggests that the synthetic data $\hat{y}$ is diverging from the real data distribution.
These metrics quantify the domain shift \cite{kouw2018introduction}, which as we showed undermines statistical guarantees of MLE objective.



\mypara{Experimental setup.} For GER methods, degradation in $\sdomain$ directly correlates with downstream performance evaluated on $\domain$. 
\begin{wrapfigure}{r}{0.48\textwidth}
  \centering
  \includegraphics[width=0.45\textwidth]{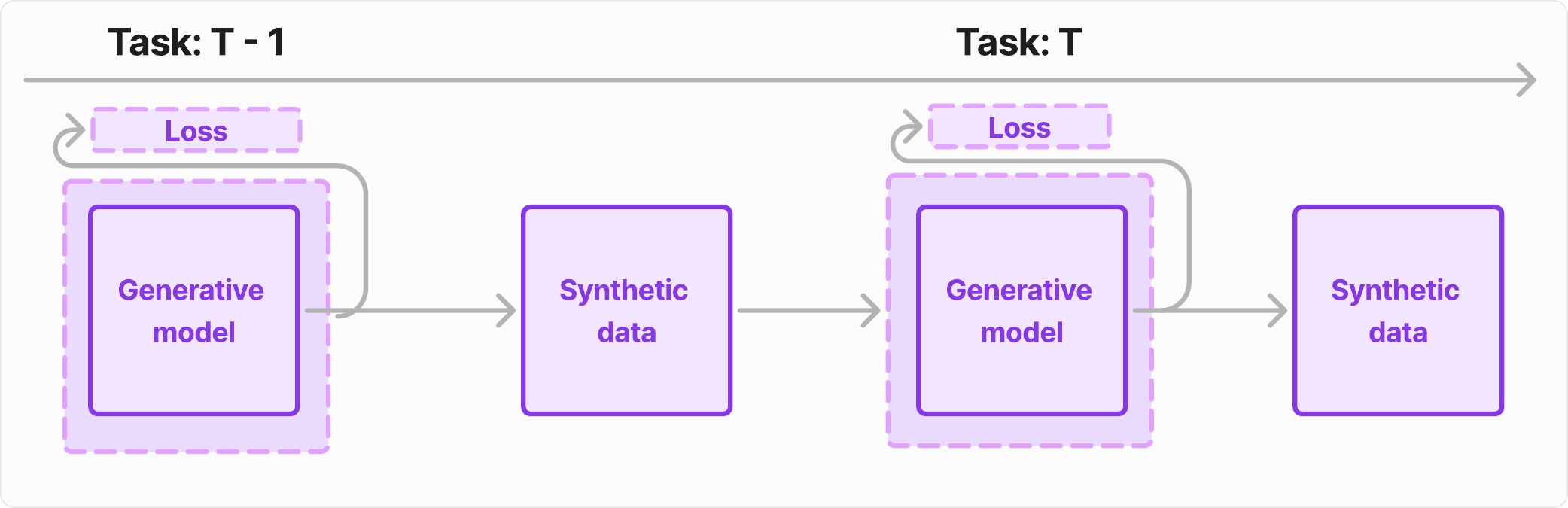}
  \caption{The generative model is sequentially trained on synthetic data sampled from the weights obtained in the previous task.}
  \label{fig:gen_bootstrap}
\end{wrapfigure}
To find an upper bound on model stability and quantify domain drift, we consider a scenario where the model is never exposed to the real domain.

The training consists of $T$ stages shown in Fig.~\ref{fig:gen_bootstrap}. In the first stage, the generative model is trained on the entire real dataset $\domain_{\text{dataset}} = \bigcup_{i=1}^T \domain_i$. Subsequently, the real domains are replaced with synthetic domains $\sdomain_{\text{dataset}} = \bigcup_{i=1}^T \sdomain_i$, and the model is retrained on this synthetic data. 

To assess the stability of the generator we compute FID, OTDD, and classification scores. 

\begin{figure*}[t]
  \centering

  \begin{subfigure}[t]{0.48\textwidth}
    \centering
    \includegraphics[width=\textwidth]{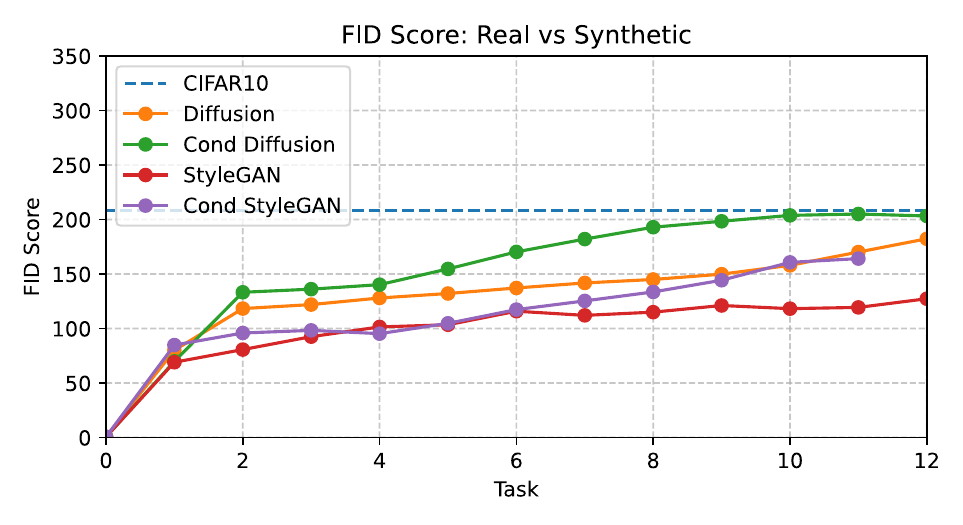}
    \caption{
      FID score between the generated dataset and the Flower102 dataset.
    }
    \label{fig:vanila-fid-collaps}
  \end{subfigure}
  \hfill
  \begin{subfigure}[t]{0.48\textwidth}
    \centering
    \includegraphics[width=\textwidth]{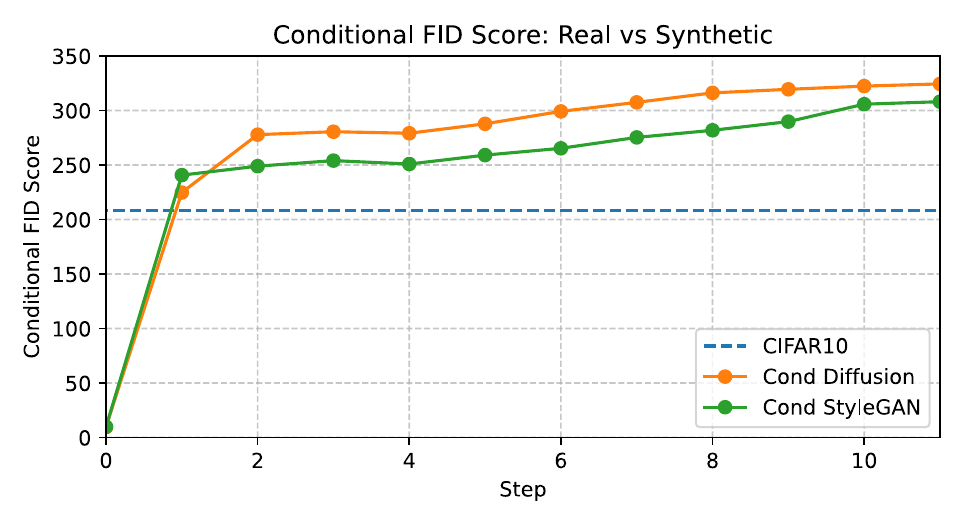}
    \caption{
      CFID score between the conditional synthetic dataset and the Flower102 dataset.
    }
    \label{fig:cond-fid-collaps}
  \end{subfigure}
   \begin{subfigure}[t]{0.47\textwidth}
    \centering
    \includegraphics[width=\linewidth]{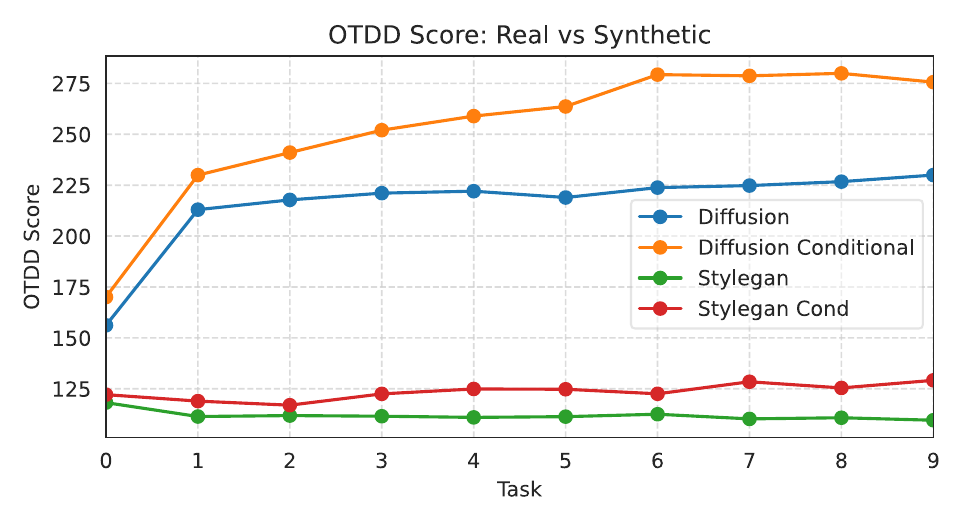}
    \caption{OTDD score between the generated synthetic dataset and the Flower102 dataset.
    }
    \label{fig:otdd-collaps}
  \end{subfigure}
  \hfill
  \begin{subfigure}[t]{0.47\textwidth}
    \centering
    \includegraphics[width=\linewidth]{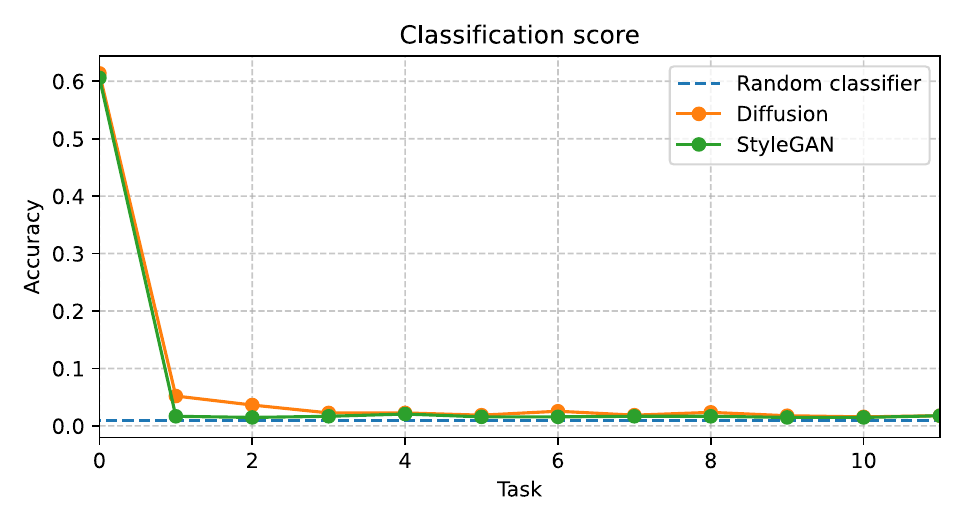}
    \caption{Classification accuracy of ResNet18 trained on synthetic data and validated on Flower102.
    }
    \label{fig:classification-collaps}
  \end{subfigure}
  \caption{
    Comparison of FID, CFID, OTDD, and classification accuracy between synthetic data and the original Flower102 dataset~\cite{Nilsback08} measured at the end of each task. 
    The baseline line in (a) and (b) in blue shows the FID / CFID score between Flower102 and CIFAR10.
  }
  \label{fig:fid-collaps}
\end{figure*}

In our experiments, 
we use Flower102~\cite{Nilsback08} as our initial real dataset $\domain_{\text{dataset}}$ and train two class-unconditional generative models: Denoising Diffusion Probabilistic Model (DDPM)~\cite{ho2020denoising} and StyleGANv2~\cite{karras2019analyzing}.
Furthermore, we train a class-conditioned variant of DDPM where, at every denoising step, we concatenate a learnable class embedding to the time embedding. For the conditioned StyleGANv2, we follow~\cite{oeldorf2019loganv2}, concatenating a learnable class embedding with the latent input for the mapping network. 
We train each model for several stages. At the end of each stage $t$, we generate synthetic samples $\{\sdomain_i^t\}_{i=1}^{N_t}$ equal in size to Flower102. These samples are used to compute our evaluation metrics and serve as the training input for the next stage.

\mypara{Experimental findings.} We observe that over time FID and CFID scores increase (Figs.~\ref{fig:vanila-fid-collaps} and \ref{fig:cond-fid-collaps}), indicating that the generated images increasingly differ from the original dataset in the image and latent space. We include the FID score between Flower102 and CIFAR-10 as a baseline (blue line) to flag that the synthetic samples eventually are as different from the initial dataset as CIFAR-10 is. We provide visual examples in the Supplementary.

Fig.~\ref{fig:otdd-collaps} shows similar trends for OTDD. Interestingly, over the course of the training process, the sampled datasets from the StyleGAN family remain stable, suggesting that the latent distribution of the GAN model does not change significantly, even if the sample quality decreases.
Fig.~\ref{fig:classification-collaps} shows that the domain drift between the distribution of original labels $y$ and synthetic labels $\hat{y}$ occurs faster than that between $x$ and $\hat{x}$. The classification performance almost collapses to the lower boundary of a random classifier within a few steps, implying that both model families are unstable in modelling the distribution of labels.
%
The results confirm that the statistical reliability of the weights decreases significantly over time.

\subsection{GER model stability -- qualitative analysis of the latent space}\label{sec:latent_space_analysis}
In this section, we analyse the latent spaces of two GER methods: GAN-Memory~\cite{cong2020gan} and DDGR~\cite{gao2023ddgr}. These methods represent distinct training procedures. GAN-Memory adopts a separate training strategy, whereas DDGR follows a joint training approach. We selected these methods because they are among the few capable of operating on large-scale datasets, including ImageNet~\cite{deng2009imagenet}, CelebA~\cite{liu2015faceattributes}, and Flower102~\cite{Nilsback08}. We show that even though these methods were specifically designed to work with synthetic data, the latent discrepancy between the real domain $\domain$ and the synthetic domain $\sdomain$ remains significant.

\begin{figure*}
  \centering
  \begin{subfigure}{0.3\linewidth}
    \includegraphics[width=\linewidth]{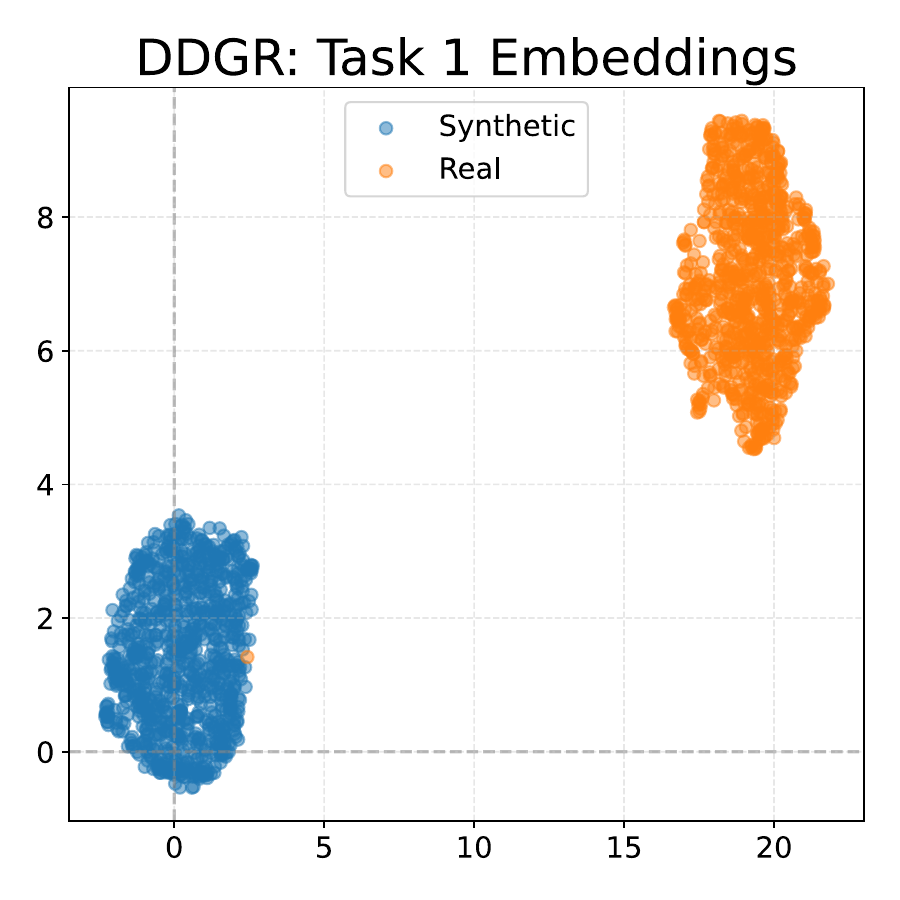}
    \caption{DDGR classifier embeddings projected with UMAP.}
    \label{fig:ddgr-embs-projections}
  \end{subfigure}
  \hfill
  \begin{subfigure}{0.3\linewidth}
    \includegraphics[width=\linewidth]{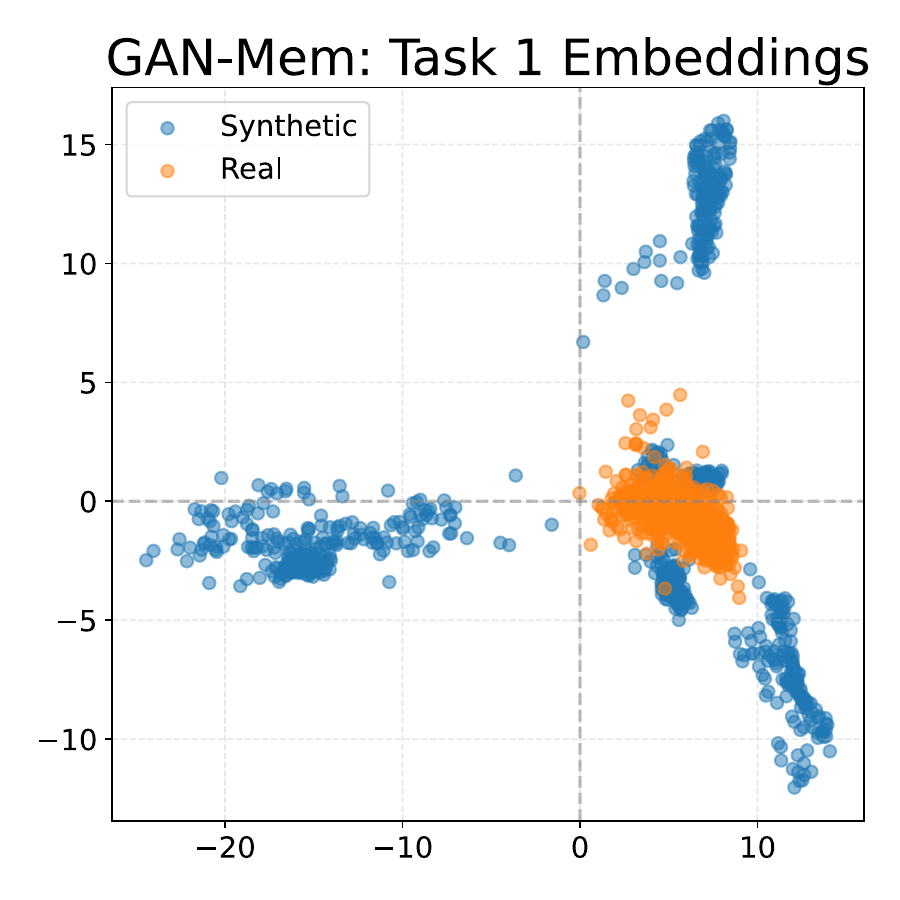}
    \caption{GAN memory classifier trained on synthetic data $\sdomain$.}
    \label{fig:ganmem-embs-synth-clf}
  \end{subfigure}
  \hfill
  \begin{subfigure}{0.3\linewidth}
    \includegraphics[width=\linewidth]{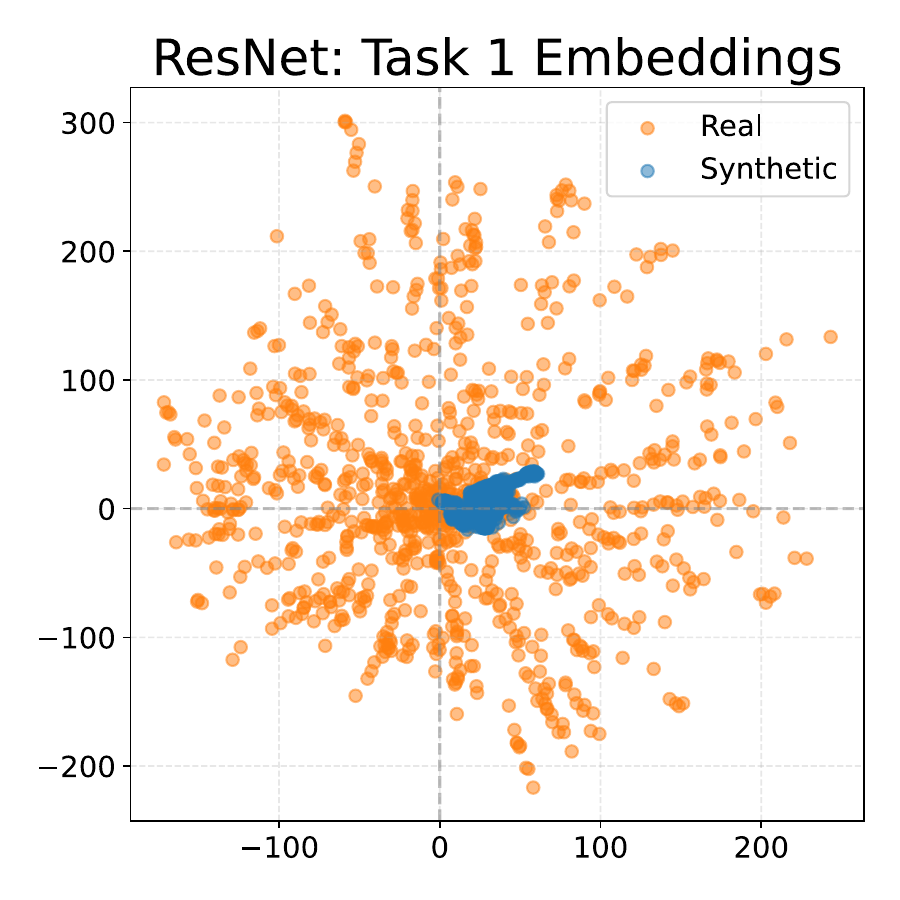}
    \caption{GAN memory classifier trained on real data $\domain$.}
    \label{fig:ganmem-embs-real-clf}
  \end{subfigure}
  \caption{Visualisation of embeddings for DDGR and GAN memory classifiers for real (orange) and synthetic domains (blue). See the Supplementary for additional tasks.}
  \label{fig:combined-latent-space-embeddings}
\end{figure*}

\mypara{Deep Diffusion-based Generative Replay.} Following the procedure from the original paper, we train DDGR on CIFAR-100 \cite{krizhevsky2009learning} split into five tasks, and generate a set of images at the end of each training phase. We then fit a UMAP \cite{McInnes2018} model to the concatenation of classifier embeddings for real task data and generated synthetic samples. Fig.~\ref{fig:ddgr-embs-projections} illustrates the embedding space of the classifier, where one can observe that starting from the first phase, embeddings for real and synthetic data are separated on the embedding manifold.
Additionally, we linearly probe~\cite{alain2016understanding} the embeddings to distinguish real from synthetic data. We achieve perfect or near-perfect accuracy at each task ($\geq 99.9\%$). This provides further strong evidence that the model separates the domains.

\mypara{Generative memory buffer.}
GAN Memory \cite{cong2020gan} uses a GAN~\cite{goodfellow2014generative} whose outputs supervise the classifier. 
We use their training procedure, separating the Flower102 dataset into 6 tasks, each containing 17 classes. We train GAN Memory to reproduce samples from tasks. To investigate the latent structure of the learned model, we train classifiers (Resnet18) for each task on GAN-generated samples. GAN Memory uses a conditional GAN, the synthetic domains $\sdomain_i$ are obtained with data samples $\hat{x}$ and their respective labels $\hat{y}$.
For visualisation, we use the dimensionality bottleneck from~\cite{vaze2021openset} (details in Supplementary) which projects the latent space into two dimensions. We observe in Fig. \ref{fig:ganmem-embs-synth-clf} that the real domain is projected close to the origin, indicating that it is out of distribution for this model.

We also consider the reversed scenario, where the classifier (ResNet50) is trained on the original dataset $\domain_{\text{dataset}}$ (Fig.~\ref{fig:ganmem-embs-real-clf}). We observe the same behaviour: the synthetic domains not only differ significantly in latent space shape from the real domain but are also generally closer to the centre of coordinates than any part of the real star-shaped latent space.

\section{Conclusion}
In this work, we analysed statistical properties of generative models in the context of continual learning methods, such as Generative Experience Replay.
We found that collapse is an inevitable property, both with mild theoretical assumptions and verified it empirically in standard CL benchmarks. This points to fundamental issues with GER as a tool for CL, which must be resolved to achieve the goal of maintaining performance through repeated training and generation iterations.
Furthermore, our findings point towards a necessary critical reevaluation of the growing proportion of synthetic data in the internet data corpus used for the continual training of models.

\bibliographystyle{splncs04}
\bibliography{egbib}

\newpage
\appendix 

\section{Latent space structure}
\setcounter{page}{1}

In this section, we provide details about the experiment used to illustrate Equations~\ref{eq:mle-bias} and \ref{eq:mle-variance}. Furthermore, we describe the motivation for the conditional FID score (CFID) proposed in Section~\ref{sec:quantitative_analysis_gen}.

\subsection{Wasserstein GAN divergence}
We start with describing the experiment used to visualise increasing variance in Section~\ref{sec:train_objective}. The initial dataset was obtained by sampling from a mixture of five 2D Gaussians. The centres of these Gaussians are located at:
$$[(0, 0), (3, 3), (-3, 3), (-3, -3), (3, -3)]$$
Their standard deviations are 
$$[(1.2, 1.0), (1.1, 0.9), (1.0, 1.2), (1.0, 1.1), (0.9, 1.0)]$$ respectively.
This configuration yields the initial density depicted in Figure~\ref{fig:density_phase_0} in the main body of the paper.

As a generative model, we choose Wasserstein GAN (WGAN) \cite{arjovsky2017wasserstein}. The generator of the GAN is a stack of three linear layers with dimensions $ z_{\text{dim}} \times 32 \times 64 \times 2$, with $z_{\text{dim}} = 2$, separated by ReLU activation functions \cite{agarap2018deep}. The discriminator is a stack of three linear layers with dimensions $ 2 \times 64 \times 32 \times 1$, separated by LeakyReLU activation functions \cite{xu2015empirical}.

We start the procedure by training WGAN on 100,000 samples from the GMM (Task 1, Figure \ref{fig:inceptionnet_density_all_classes}). After successful training, we sample 5,000 samples from the GAN (20 times smaller than the GMM dataset), simulating the undersampling problem described in Section~\ref{sec:stat_errors}, and retrained the WGAN on its own generations (Task 2, Figure \ref{fig:inceptionnet_density_all_classes}). The process then repeats 10 times; each new W-GAN model is trained on the generations of the previous WGAN model.

\begin{figure*}
  \centering
  \includegraphics[width=\linewidth]{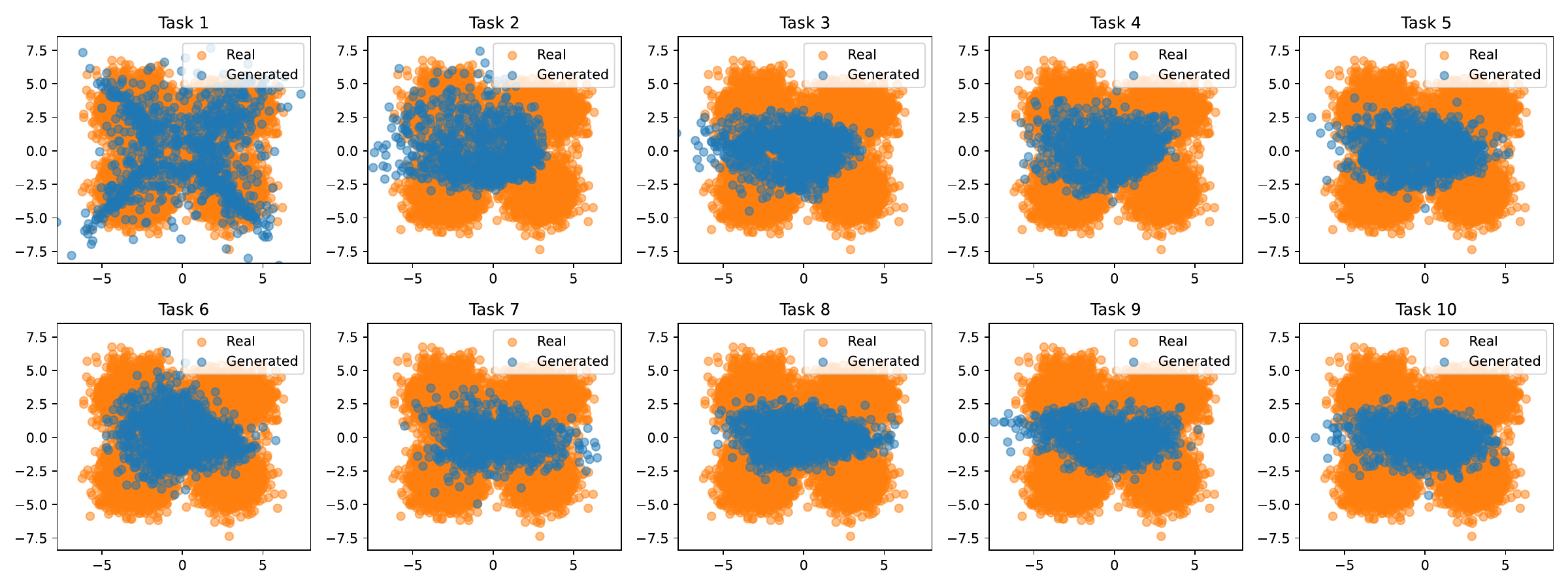}
  \caption{Visualisation of GMM-generated (real dataset) and WGAN-generated samples at each task. Orange points are samples from the mixture of five Gaussians, and blue points are generations from the WGAN obtained at the end of each task.}
  \label{fig:wgan-vs-gmm-all-tasks}
\end{figure*}

Figure~\ref{fig:wgan-vs-gmm-all-tasks} depicts the sample distributions from the GAN and GMM models obtained at the end of each task. We observe that over time, the distributions of the GMM and WGAN visually diverge, leading to increasing bias (Eq.~\ref{eq:mle-bias}) and variance (Eq.~\ref{eq:mle-variance}) of the Maximum Likelihood Estimator (MLE).

\subsection{Relation between MLE and FID scores}
\begin{figure}
  \centering
  \includegraphics[width=0.8\linewidth]{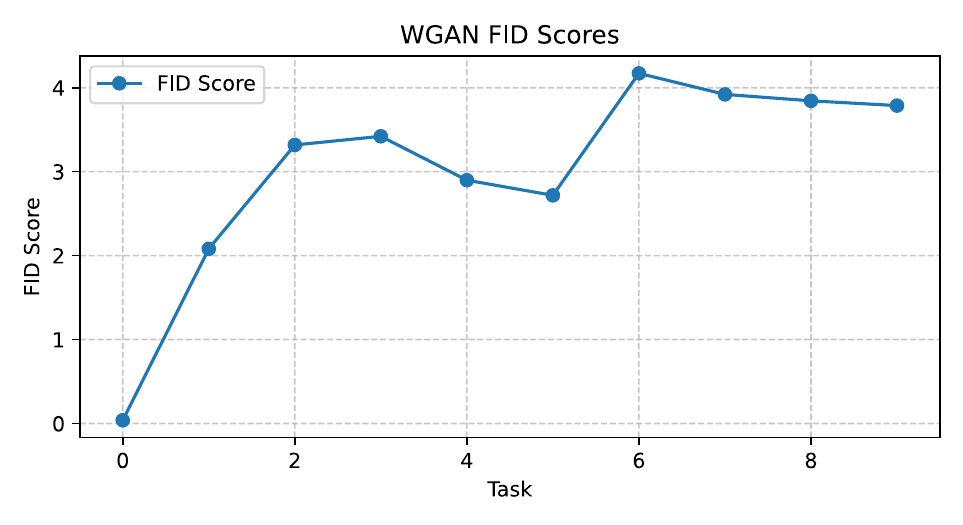}
  \caption{FID Score between WGAN samples and GMM samples for each task.}
  \label{fig:2d_gan_fid_scores}
\end{figure}

In the main paper, we frequently mention that divergence, as measured by the Fréchet Inception Distance (FID) score, is connected to increasing bias (Eq.~\ref{eq:mle-bias}) and variance (Eq.~\ref{eq:mle-variance}) in the Maximum Likelihood Estimator (MLE). This theoretical connection arises from the fact that the FID score represents the Fréchet Distance between two probability distributions. Thus, as the FID score increases, it indicates a greater divergence between the synthetic and real data distributions, which in turn contributes to higher bias and variance in parameter estimation using the MLE. For illustrative purposes, we also provide Figure~\ref{fig:2d_gan_fid_scores}, which shows that over time, the FID score between the distribution generated by the Wasserstein GAN (WGAN) and the original Gaussian Mixture Model (GMM) distribution increases.

\subsection{InceptionNet latent space}
Our proposed conditional FID score is motivated by the structure of the latent space in InceptionNet \cite{szegedy2015rethinking}. To visualise the output of the encoder directly and avoid dimensionality reduction techniques (e.g.\ UMAP \cite{McInnes2018}),\cite{vaze2021openset} suggest creating an artificial dimensionality bottleneck by reducing the encoder activation space to two dimensions. Here, the encoder refers to all layers before the last fully connected layer. When such an encoder is trained, it learns to project the dataset into a `star-shaped' latent space.

\begin{figure*}[t]
  \centering

  \begin{subfigure}[b]{0.32\linewidth}
    \centering
    \includegraphics[width=0.95\linewidth]{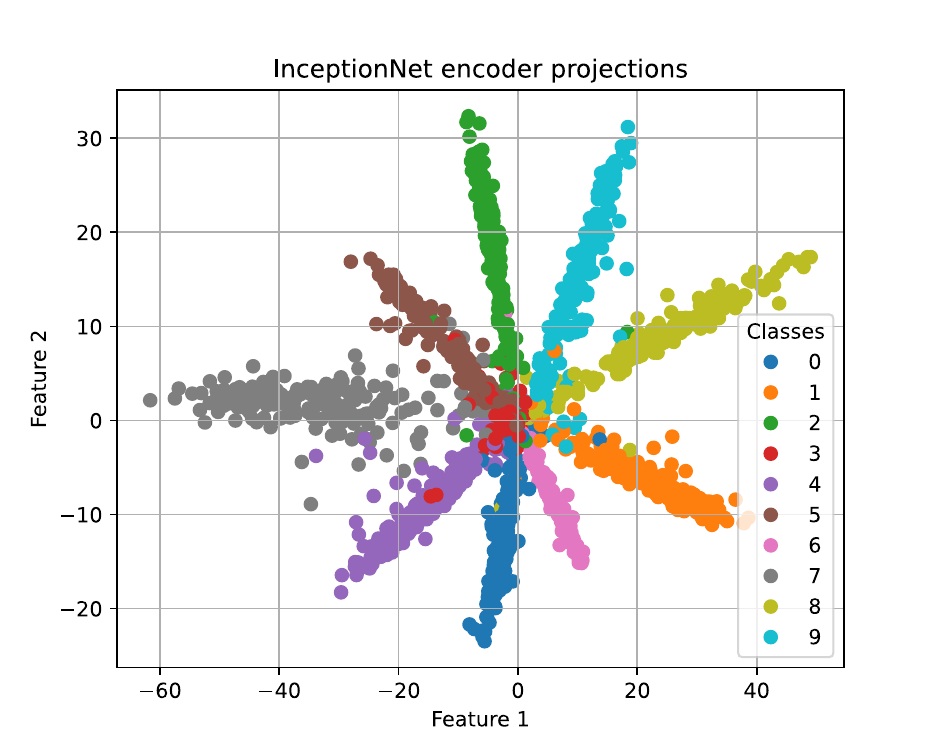}
  \caption{InceptionNet encoder projections of the CIFAR-10 dataset \cite{krizhevsky2009learning} with a dimensionality bottleneck of 2.}
    \label{fig:inceptionnet_2d_embs}
  \end{subfigure}
  \hfill
  \begin{subfigure}[b]{0.32\linewidth}
    \centering
    \includegraphics[width=0.95\linewidth]{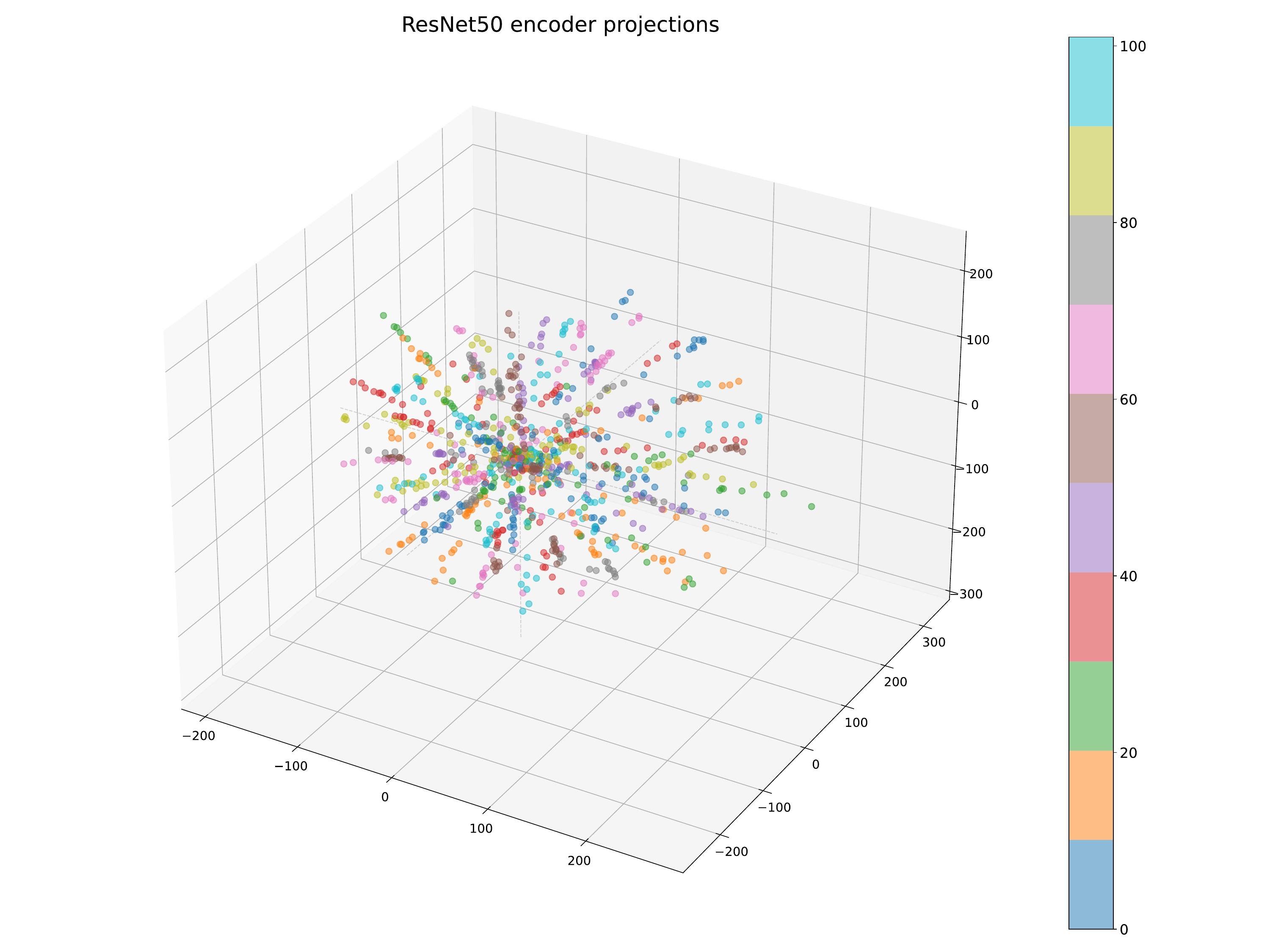}
  \caption{ResNet-50 \cite{he2015deep} encoder projections of the CIFAR-100 dataset \cite{krizhevsky2009learning} with a dimensionality bottleneck of 3.}
    \label{fig:resnet_3d_embs}
  \end{subfigure}
  \hfill
  \begin{subfigure}[b]{0.32\linewidth}
    \centering
    \includegraphics[width=0.95\linewidth]{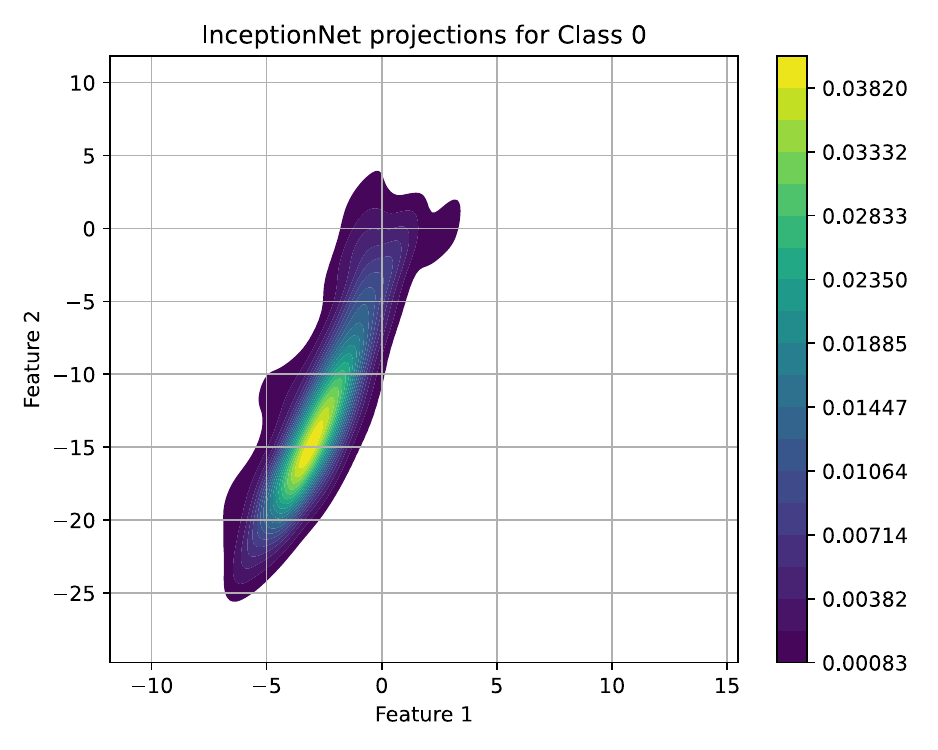}
  \caption{Kernel Density Estimations (KDE) \cite{scott2015multivariate} for class 0 computed on InceptionNet encoder outputs.}
    \label{fig:inceptionnet_density_class_0}
  \end{subfigure}

  \caption{Visualisation of encoder outputs from different models.}
  \label{fig:combined_embeddings}
\end{figure*}

To illustrate this finding, we set the dimensionality bottleneck to 2 and train an InceptionNet v3 on the CIFAR-10 dataset \cite{krizhevsky2009learning} until it achieves 92\% accuracy — a performance comparable to standard training. We then extract the encoder activations for 2,000 randomly selected samples from the validation dataset. Figure \ref{fig:inceptionnet_2d_embs} depicts the `star-shaped' latent space learned by InceptionNet v3 in $\mathcal{R}^{2}$.

\cite{vaze2021openset} argue that such a latent space structure arises because, in the cross-entropy objective for a single sample in the batch (Equation \ref{eq:cross-entropy}), the last linear layer performs a dot product between the encoder output $\Phi_\theta(\mathbf{x}_i)$ and the rows $\mathbf{w}_j$ of the matrix $W$ that parameterizes the linear layer:  
\begin{align}
\label{eq:cross-entropy}
\mathcal{L}_i(\boldsymbol{\theta}, \mathbf{W}) &= -\hat{y}_{i,c} + \log\left(\sum_{j=1}^{C} \exp(\hat{y}_{i,j})\right) \\
    &= -\mathbf{w}_c \cdot \Phi_\theta(\mathbf{x}_i) + \log\left(\sum_{j=1}^{C} \exp(\mathbf{w}_j \cdot \Phi_\theta(\mathbf{x}_i))\right).
\end{align}
This formulation encourages the encoder outputs to align with their corresponding class weight vectors, due to dot products $\mathbf{w}_j \cdot \Phi_\theta(\mathbf{x}_i)$, resulting in a `star-shaped' latent space where each `wing' corresponds to a different class.

Traditional training of InceptionNet v3 would correspond to a dimensionality bottleneck equal to $z_{\text{dim}}$, resulting in a $k$-dimensional `star-shaped' latent space. For example, Figure \ref{fig:resnet_3d_embs} shows the encoder projections of a ResNet-50 \cite{he2015deep} trained on the CIFAR-100 dataset \cite{krizhevsky2009learning} with a dimensionality bottleneck of 3.

We observe that the distribution of the samples in Figure \ref{fig:inceptionnet_2d_embs} deviates significantly from a normal distribution, which is a key assumption for the FID score \cite{heusel2017gans}. Since the sample distribution is not normal, this can lead to inaccuracies in the FID measurement. We propose to address this by considering the `wings' of the star-shaped distribution separately when calculating the FID score.

\begin{figure*}
  \centering
  \includegraphics[width=\linewidth]{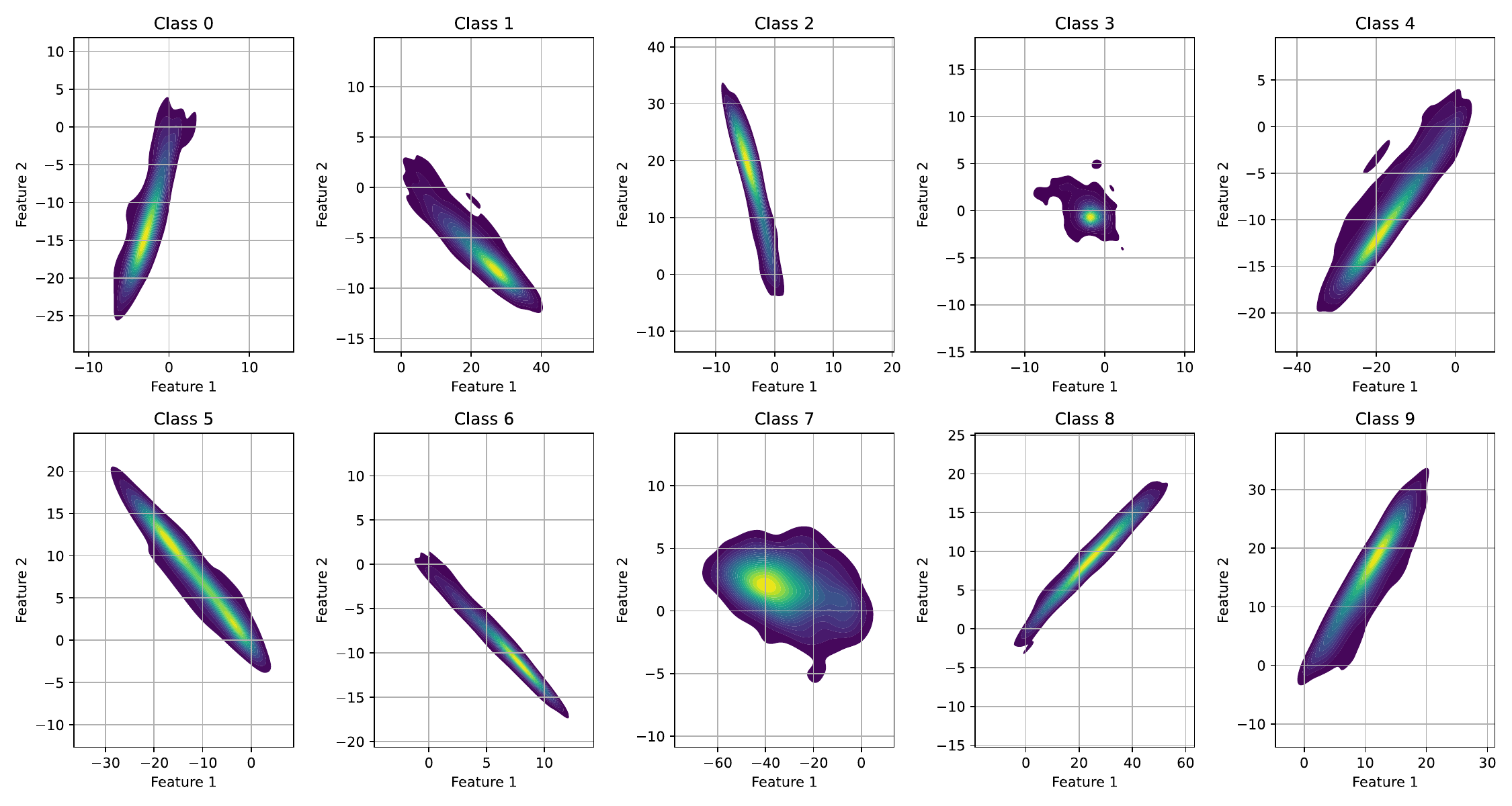}
  \caption{Kernel Density Estimations (KDE) \cite{scott2015multivariate} for all classes computed on InceptionNet encoder outputs.}
  \label{fig:inceptionnet_density_all_classes}
\end{figure*}

Figure \ref{fig:inceptionnet_density_class_0} illustrates the Kernel Density Estimation (KDE) for class 0, showing that class conditional distribution $\Phi_\theta(\mathbf{x}_i) | y_i$ is closer to Gaussian distribution. Figure \ref{fig:inceptionnet_density_all_classes} depicts KDE estimations for the remaining classes.

By calculating the FID score for each class separately—effectively treating each `wing' independently — we can obtain a more accurate assessment of the generative model's performance. This approach forms the basis of our proposed conditional FID score.

\subsection{Conditional FID score}

The FID score uses a closed-form solution of the Fréchet Distance $d_{F}$ between two Gaussian distributions $\mathcal{N}(\mu_1, \Sigma_1)$ and $\mathcal{N}(\mu_2, \Sigma_2)$:
\begin{align}
\label{eq:fid}
    &d_{F}(\mathcal N(\mu_1, \Sigma_1), \mathcal N(\mu_2, \Sigma_2))^2  = 
     = \lVert \mu_1 - \mu_2 \rVert^2_2 + \operatorname{tr}\left(\Sigma_1 + \Sigma_2-2\left(\Sigma_1 \Sigma_2  \right)^\frac{1}{2} \right).
\end{align}
The parameters $\mu_1$, $\Sigma_1$, $\mu_2$, and $\Sigma_2$ are estimated empirically from the activations of the InceptionNet v3 encoder:
\begin{align}
    \mu_1  &= \frac{1}{N} \sum_{j=1}^N \Phi_\theta(x_{1j}), \\
    \Sigma_1  &= \frac{1}{N-1} \sum_{j=1}^N(\Phi_\theta(x_{1j}) - \mu_1 )^2, \\
    \mu_2  &= \frac{1}{N} \sum_{j=1}^N \Phi_\theta(x_{2j}), \\
    \Sigma_2  &= \frac{1}{N-1} \sum_{j=1}^N(\Phi_\theta(x_{2j}) - \mu_2 )^2,
\end{align}
where $\Phi_\theta(\cdot)$ denotes the InceptionNet v3 encoder, and $x_{1j}$ and $x_{2j}$ are samples from datasets $X_1$ and $X_2$, respectively, between which we compute the FID score. 

In practice, this estimation assumes that the activations of the InceptionNet v3 encoder on our datasets $X_1$ and $X_2$ follow Gaussian distributions. However, this is a strong assumption that may not hold due to the `star-shaped' latent space structure. 

To mitigate this issue, we propose considering the expected FID score computed between the conditional distributions $X_1 | Y = c$ and $X_2 | Y = c$, where $Y$ represents the class labels. We start by estimating the mean and covariance of the conditional distributions:
\begin{align}
    \mu_{1,c} &= \frac{1}{N_c} \sum_{j=1}^N \Phi_\theta(x_{1j}) * \mathcal{I}[y_i = c], \\
    \Sigma_{1,c}  &= \frac{1}{N_c-1} \sum_{j=1}^N(\Phi_\theta(\mathbf{x}_i)) - \mu )^2 * \mathcal{I}[y_i = c],
\end{align}
where $\mathcal{I}[y_j = c]$ is an indicator function that equals 1 when $y_j = c$ and 0 otherwise, and $N_c$ is the number of samples in dataset $X_1$ with class label $c$. We compute $\mu_{2,c}$ and $\Sigma_{2,c}$ similarly for dataset $X_2$.

By estimating the mean and covariance for all conditional distributions, we define our Conditional Fréchet Inception Distance as the expectation over the variable $Y$ that takes $k$ realisations / classes:
\begin{align}
    \mean{Y}{d_F(X_1|Y , X_2|Y)^2} &= 
     \frac{1}{k} \sum_{c=1}^k d_F(X_1 |\, Y = c, X_2 |\, Y = c)^2 \nonumber \\
    &= \frac{1}{k} \sum_{c=1}^k \lVert \mu_{1,c} - \mu_{2,c} \rVert^2_2 + \operatorname{tr}\left(\Sigma_{1,c} + \Sigma_{2,c} -2\left(\Sigma_{1,c} \Sigma_{2,c}  \right)^\frac{1}{2} \right) \nonumber.
\end{align}
This formula calculates the average Fréchet Distance between subspaces of the activation space, thereby better capturing changes in the `star-shaped' latent spaces of classifier models.

\section{DDGR and GAN Memory latent spaces}
In this section, we provide additional plots related to the experiments in Section \ref{sec:latent_space_analysis}.

\subsection{DDGR}

\begin{figure*}[t]
  \centering
  \includegraphics[width=\linewidth]{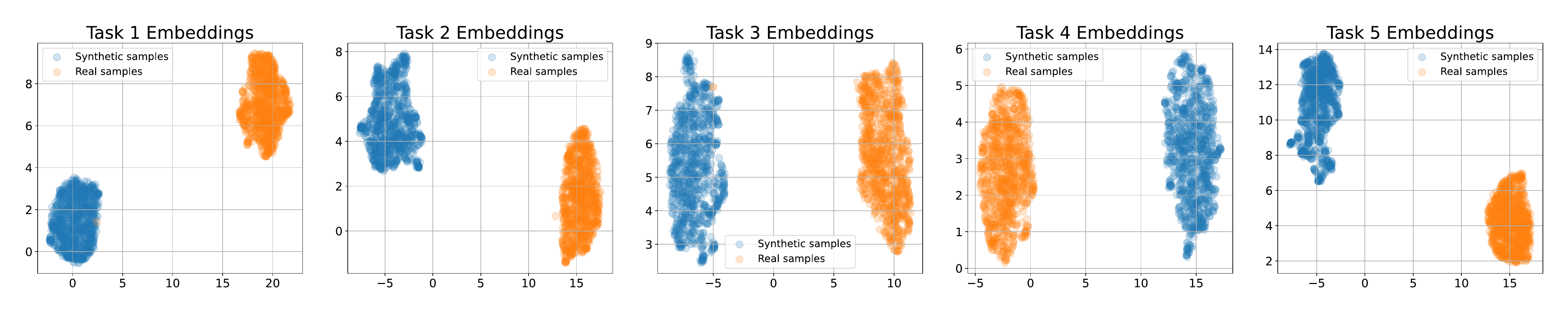}
  \caption{Visualisation of the latent space of the DDGR classifier computed with UMAP at the end of every task. We first project generated or real images to the classifier latent space and then use UMAP to project them to $\mathbb{R}^2$.}
  \label{fig:ddgr-emb-all-tasks}
\end{figure*}

For our experiments, we used the official code provided by the authors at \url{https://github.com/xiaocangshengGR/DDGR}. Without any changes, we ran the code on the CIFAR-100 dataset, which is split into five tasks, according to the procedure in \cite{gao2023ddgr}, and then investigated the produced artefacts to visualise the latent space of the model.

The training generates five checkpoints for classifiers and five checkpoints for generators, one for each task. We use them to generate synthetic datasets equal in size to the CIFAR-100 dataset and then fit a UMAP model on both real and synthetic data. We used UMAP here because we were not able to use the same dimension reduction techniques for latent space visualisation as described in Supplementary A, since they made DDGR training unstable.
 
Figure \ref{fig:ddgr-emb-all-tasks} depicts UMAP projections for both real and synthetic samples, coloured respectively. One can see that starting from the first task, when the generator according to the DDGR training procedure was only exposed to real samples, the classifier model already separates real images from generated ones. This behaviour persists across new tasks. The classifier consistently projects synthetic data and real data to two different distributions, and more importantly, according to the UMAP projections, to two non-intersecting distributions.

We also trained a linear head on the original-sized embedding space of the frozen classifier and were able to achieve perfect or near-perfect validation accuracy, depending on the task, suggesting that embeddings of synthetic and real data are linearly separable.

\subsection{GAN Memory}

\begin{figure*}[t]
  \centering
  \includegraphics[width=\linewidth]{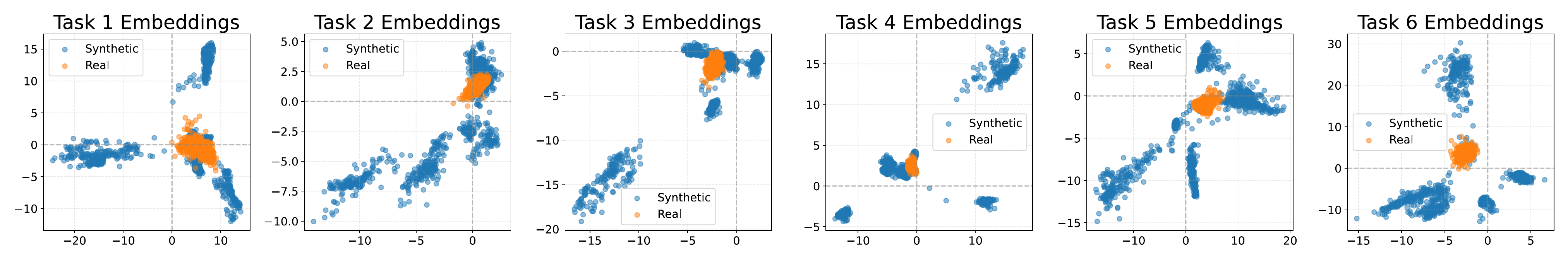}
  \caption{Visualisation of ResNet-18 encoders outputs for every task. For each task projection, we use the corresponding model that was trained on synthetic data reproducing the task dataset. Since we use a dimensionality bottleneck, all the outputs belong to $\mathbb{R}^2$.}
  \label{fig:ganmem-emb-2d-all-tasks}
\end{figure*}

\begin{figure*}[t]
  \centering
  \includegraphics[width=\linewidth]{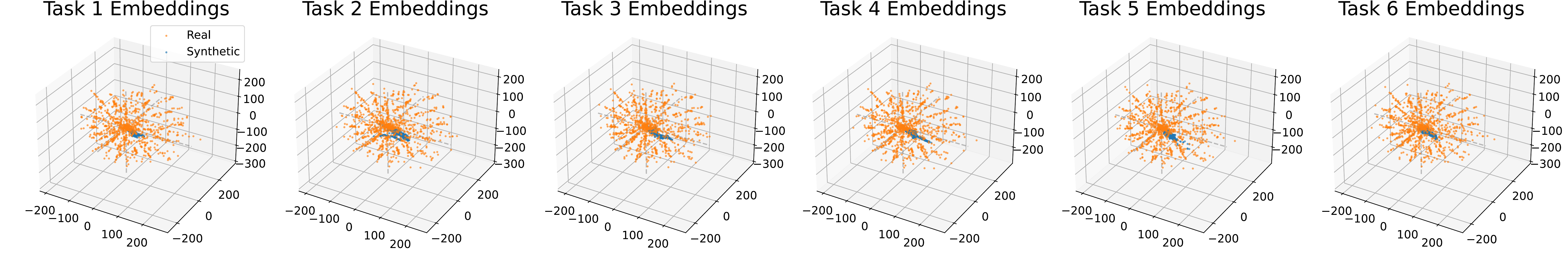}
  \caption{Visualisation of ResNet-50 outputs for every task. Orange-coloured points are projections of real samples, and blue-coloured points are projections of synthetic samples from the synthetic dataset produced in task $k$.}
  \label{fig:ganmem-emb-3d-all-tasks}
\end{figure*}

For our experiments, we used the official code provided by the authors at \url{https://github.com/MiaoyunZhao/GANmemory_LifelongLearning}. Following the procedure described in the original paper \cite{cong2020gan}, we split Flower102 into six disjoint tasks, each containing 17 classes, where each class appears only in one task. We then train six different conditional GAN models that are later incorporated together as described in \cite{cong2020gan}.

Once all six GAN models are trained and incorporated, we generate six synthetic datasets, each containing an equal number of elements and classes corresponding to each task. We then train six ResNet-18 models on these datasets and use a two-dimensional bottleneck as in Supplementary A to visualize the latent space. Figure \ref{fig:ganmem-emb-2d-all-tasks} depicts embeddings for both samples from the real Flower102 dataset and the synthetic datasets. We observe that, for all tasks, real samples are closer to the origin of the coordinate system, which is a sign of out-of-domain data according to \cite{vaze2021openset}, \cite{ranjan2017l2}.

To further demonstrate that the generated synthetic dataset is different from the real Flower102, we perform a symmetric setup and train a ResNet-50 on the original Flower102 dataset, using a dimensionality bottleneck of 3. Figure \ref{fig:ganmem-emb-3d-all-tasks} depicts projections of real and synthetic datasets in the latent space of the classifier. We again observe that the activation spaces of real and synthetic datasets differ significantly.

Our experiments provide strong evidence that there is a gap in both the input and latent spaces between real and generated samples at every task.

\section{Illustration of joint and separate GER training}
\begin{figure*}[h]
  \centering

  \begin{subfigure}[t]{0.48\textwidth}
    \centering
    \includegraphics[width=\linewidth]{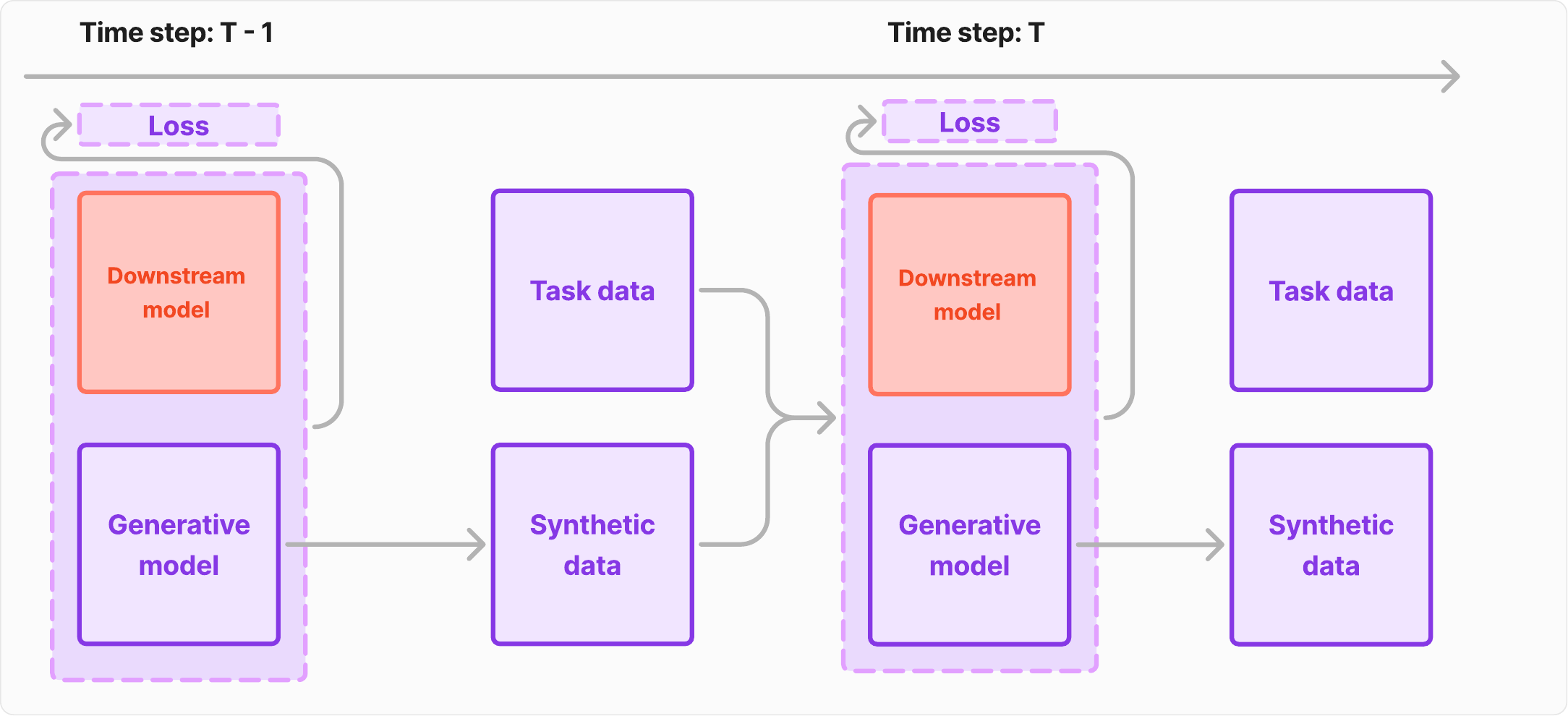}
    \caption{Joint training with shared gradients from the same batch.}
    \label{fig:ger-joined}
  \end{subfigure}
  \hfill
  \begin{subfigure}[t]{0.48\textwidth}
    \centering
    \includegraphics[width=\linewidth]{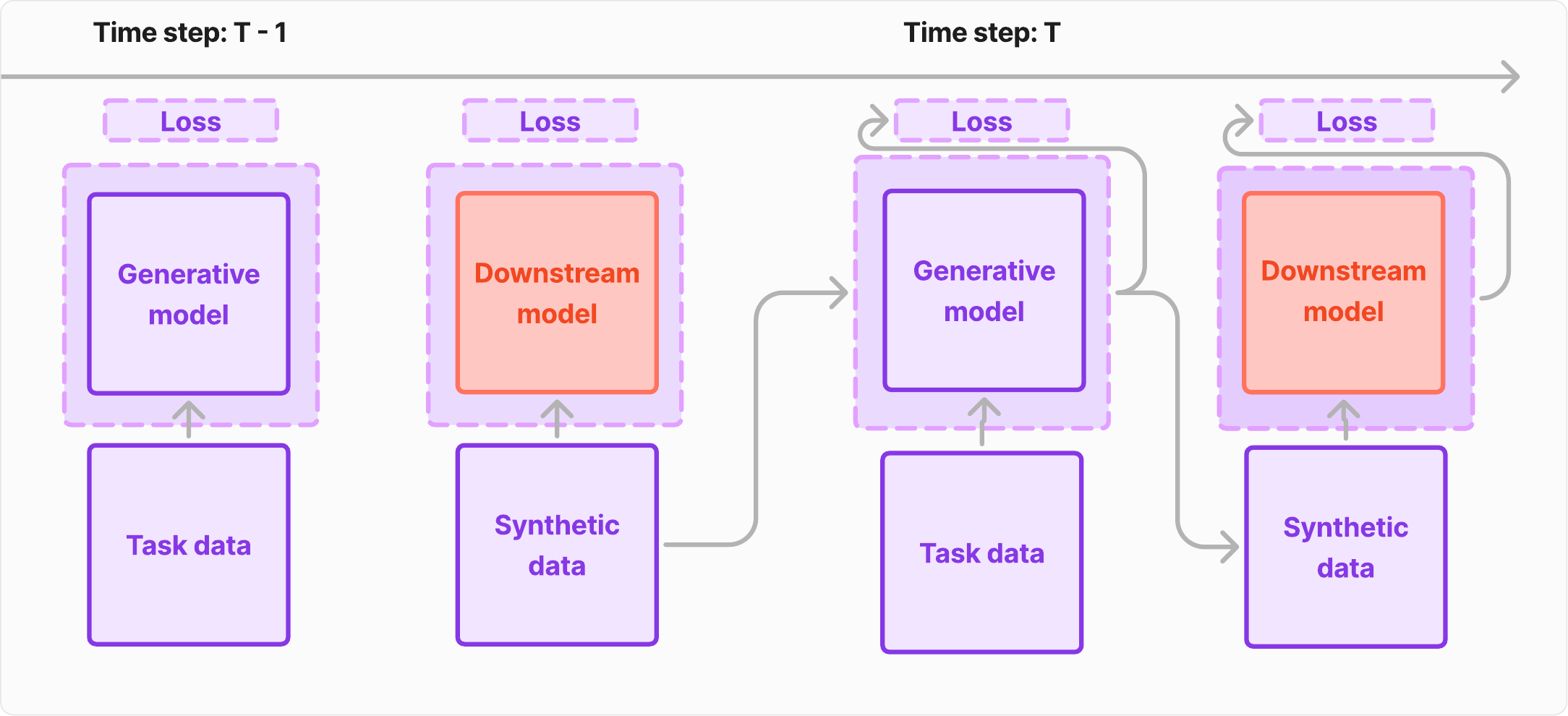}
    \caption{Separate training with different batches and steps.}
    \label{fig:ger-separate}
  \end{subfigure}

  \caption{Comparison of GER training procedures that use joint (left) and separate (right) training of the generative and downstream model.}
  \label{fig:ger_approaches}
\end{figure*}

\section{Experimental details}
In this section, we provide additional details on how to reproduce the experiments described in Section \ref{sec:quantitative_analysis_gen}.

\subsection{StyleGAN v2}
In our experiments, we used two versions of StyleGAN v2: conditional and unconditional. We use the implementation of StyleGAN v2 \cite{karras2019analyzing} from labml.ai~\cite{labml,labmlAnnotatedResearch}. The discriminator consists of six blocks with channel dimensions $[64, 128, 256, 512, 512, 512]$. The generator contains six blocks with channel dimensions $[512, 512, 256, 128, 64, 32]$. The latent space dimensionality is set to 256. The mapping network consists of eight equal linear layers \cite{karras2017progressive}. We rescale the Flower102 dataset \cite{Nilsback08} to images of size $64 \times 64$, use a batch size of 32, and the Adam optimizer \cite{kingma2014adam} with a learning rate of $1 \times 10^{-3}$ and betas $(0.0, 0.99)$. The model is trained for 150,000 steps at every task, reusing weights from the previous task if available. The other training parameters were not changed from the source code provided by labml.ai.\footnote{\url{https://nn.labml.ai/gan/stylegan/experiment.html}}

For the conditional StyleGAN v2, we follow the ideas from \cite{oeldorf2019loganv2}. Specifically, we add $k$ (the number of classes) trainable embeddings of dimensionality 128. We then concatenate the class embedding to the random input before feeding it to the mapping network. This allows the model to generate samples of a specific class. The conditional StyleGAN has the same architecture and training parameters as the unconditional version.

We refer the reader to Figure \ref{fig:cond_stylegan_task_generations} and Figure \ref{fig:stylegan_task_generations}, where we present generations from the trained models at each task.    

\subsection{DDPM}

In addition to the StyleGANs, we used two Denoising Diffusion Probabilistic Models (DDPMs). In particular, we used class-conditioned and unconditioned versions of DDPM. We utilised the implementation from the Diffusers library~\cite{von-platen-etal-2022-diffusers}. We employed the \texttt{UNet2DModel} with six downscaling blocks and six upscaling blocks. We provide the code snippet below for clarity:

\begin{lstlisting}[style=mypython, caption=Code snippet for DDPM model initialisation]
from diffusers import UNet2DModel
    
# Define model
model = UNet2DModel(
    sample_size=64,  # image resolution
    in_channels=3, 
    out_channels=3,
    layers_per_block=2, # how many ResNet layers to use per UNet block
    block_out_channels=(
        128,
        128,
        256,
        256,
        512,
        512,
    ),  # the number of output channels for each UNet block
    down_block_types=(
        "DownBlock2D",  # a regular ResNet downsampling block
        "DownBlock2D",
        "DownBlock2D",
        "DownBlock2D",
        "AttnDownBlock2D",  # a ResNet downsampling block with spatial self-attention
        "DownBlock2D",
    ),
    up_block_types=(
        "UpBlock2D",  # a regular ResNet upsampling block
        "AttnUpBlock2D",  # a ResNet upsampling block with spatial self-attention
        "UpBlock2D",
        "UpBlock2D",
        "UpBlock2D",
        "UpBlock2D",
    ),
    num_class_embeds=102 if args.conditional else None, # conditional DDPM
)
\end{lstlisting}

Similar to the StyleGAN experiments, we use the Flower102 dataset \cite{Nilsback08}, which we rescale to $64 \times 64$ pixels. We use the AdamW optimizer \cite{loshchilov2017decoupled} with a learning rate of $1 \times 10^{-4}$ and default beta parameters $(\beta_1=0.9, \beta_2=0.999)$. The batch size is set to 16, and we train for 150 epochs on each task using float16 precision. We reuse weights from the previous task, if available, when training on a new one. For learning rate scheduling, we use a cosine scheduler with 500 warmup steps.

The class-conditioned version of the DDPM shares the same architecture and training parameters. For conditioning, we add $k$ (number of classes) learnable embeddings of dimensionality matching the time embedding. These class embeddings are summed with the time embedding at every forward step to incorporate class information.

\begin{figure*}
  \centering
  \includegraphics[width=\linewidth]{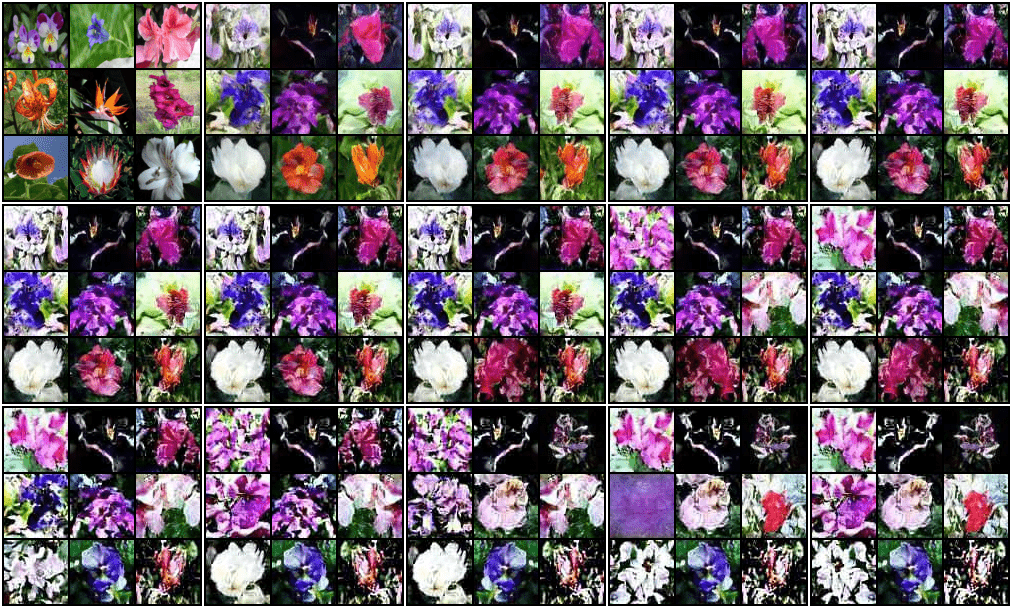}
  \caption{Random samples from the task datasets used to train the DDPM model. Task dataset samples are ordered from left to right, then from the top to bottom. The leftmost picture depicts samples from the original Flower102 dataset, the following images are generated from the DDPM model of the corresponding task.}
  \label{fig:ddpm_task_generations}
\end{figure*}

\begin{figure*}
  \centering
  \includegraphics[width=\linewidth]{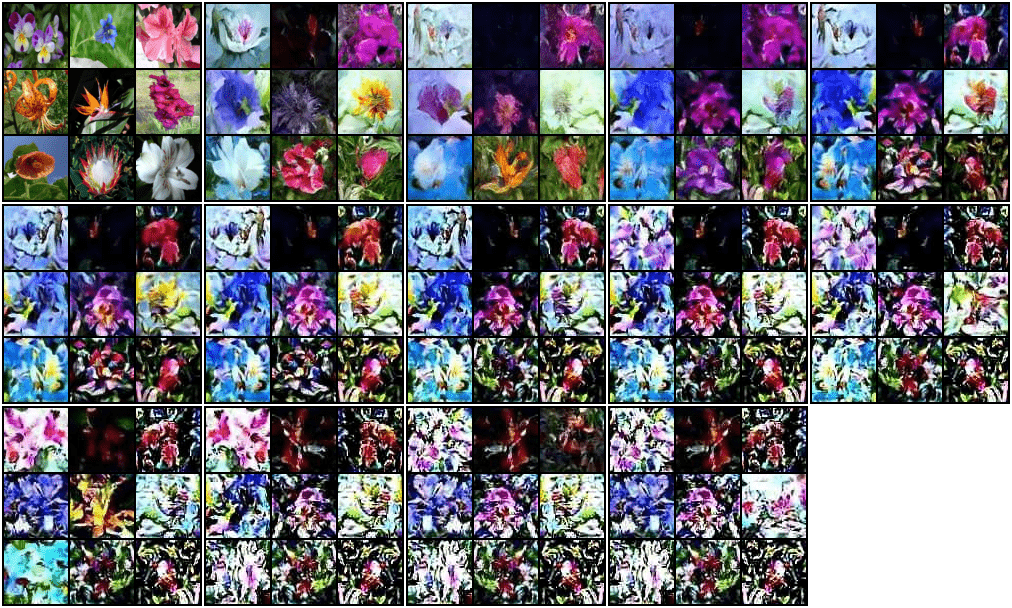}
  \caption{Random samples from the task datasets used to train the Conditional DDPM model. Task dataset samples are ordered from left to right, then from the top to bottom. The leftmost picture depicts samples from the original Flower102 dataset, the following images are generated from the Conditional DDPM model of the corresponding task.}
  \label{fig:cond_ddpm_task_generations}
\end{figure*}

We refer the reader to Figure \ref{fig:ddpm_task_generations} and Figure \ref{fig:cond_ddpm_task_generations}, where we present generations from the trained models at each task.

\section{Data Samples}
\begin{figure*}
  \centering
  \includegraphics[width=\linewidth]{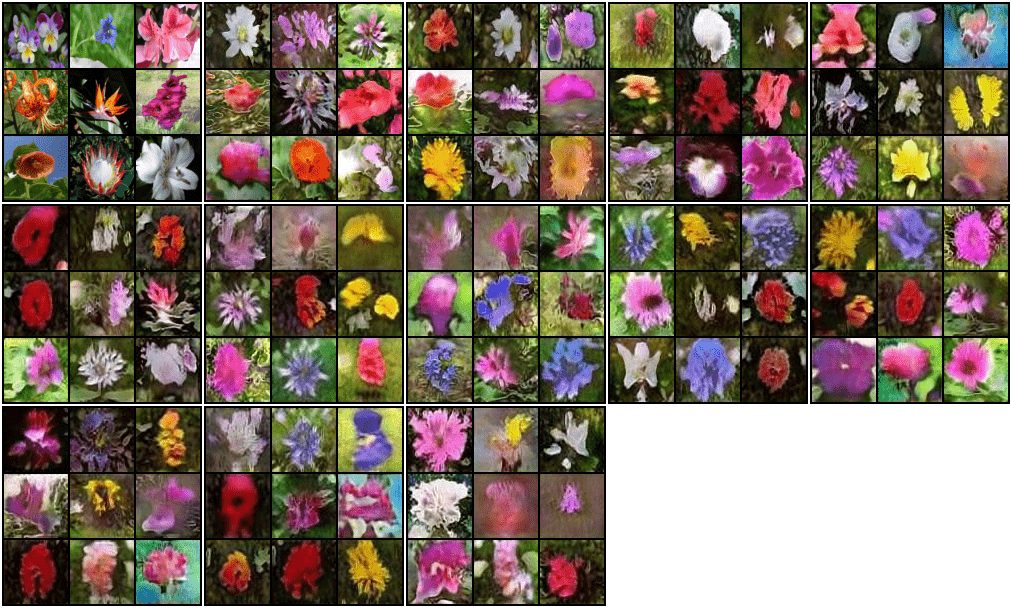}
  \caption{Random samples from the task datasets used to train the StyleGAN v2 model. Task dataset samples are ordered from left to right, then from the top to bottom. The leftmost picture depicts samples from the original Flower102 dataset, the following images are generated from the StyleGAN v2 model of the corresponding task.}
  \label{fig:stylegan_task_generations}
\end{figure*}

\begin{figure*}
  \centering
  \includegraphics[width=\linewidth]{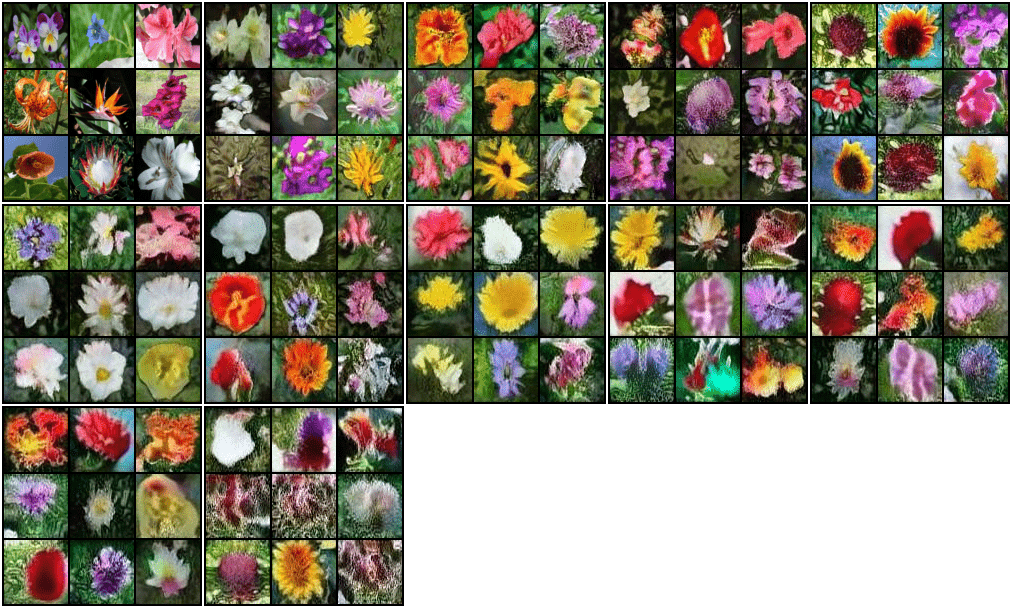}
  \caption{Random samples from the task datasets used to train the Conditional StyleGAN v2 model. Task dataset samples are ordered from left to right, then from the top to bottom. The leftmost picture depicts samples from the original Flower102 dataset, the following images are generated from the Conditional StyleGAN v2 model of the corresponding task.}
  \label{fig:cond_stylegan_task_generations}
\end{figure*}

We show examples of generated images from StyleGan v2~\cite{karras2019analyzing} and conditional SyleGAN v2~\cite{oeldorf2019loganv2} in Figures~\ref{fig:stylegan_task_generations} and \ref{fig:cond_stylegan_task_generations}.


\end{document}